\documentclass{article}
\usepackage{iclr2027_conference,times}

\iclrfinaltrue

\usepackage{amsmath,amsfonts,bm}

\def\eqref#1{equation~\ref{#1}}

\def\1{\bm{1}}

\DeclareMathAlphabet{\mathsfit}{\encodingdefault}{\sfdefault}{m}{sl}
\SetMathAlphabet{\mathsfit}{bold}{\encodingdefault}{\sfdefault}{bx}{n}

\usepackage{hyperref}
\usepackage{url}
\usepackage{enumitem}
\usepackage{booktabs}
\usepackage{graphicx}
\usepackage{wrapfig}
\usepackage{capt-of}
\usepackage{fancyvrb}
\usepackage[most]{tcolorbox}

\title{GAMEGO: Training Game-Dev Agents with Synthetic Trajectories Anchored in Real-World Assets}

\author{
Haoyue Yang$^{1,2}$, 
Jingyao Li$^{2}$, 
Zhengfan Wu$^{2}$,
Jing Liu$^{2}$,
Xuanle Zhao$^{1}$,
Kang Liu$^{1}$
\\
$^1$Institute of Automation, Chinese Academy of Sciences \\
$^2$Baidu Inc., Beijing, China\\
}

\begin{document}

\maketitle

\begin{abstract}
Recent advances in Large Language Models (LLMs) have demonstrated remarkable capabilities in web front-end execution, with browser-based game generation emerging as a particularly prominent frontier. While previous efforts frequently rely on complex multi-turn workflows or focus on static game evaluation benchmarks, this work targets direct end-to-end real-world game synthesis driven by coding agents. However, generating complex games directly from sparse user queries often forces coding agents to make underspecified assumptions, yielding incomplete mechanics, disconnected gameplay flows, and limited visual aesthetics. To resolve this issue, this paper presents GameGo, a scalable framework that systematically transforms brief game seeds into comprehensive Product Requirements Documents grounded in industry game-development practices. To retain core gameplay constraints without restricting design exploration, GameGo uses task-specific dynamic compression to maximize information density while preserving instruction following. Based on this pipeline, GameGoData is constructed with 55,060 development trajectories across 2D, 2.5D, and 3D games, alongside GameGoBench, a benchmark comprising 124 diverse game queries. Training GameGoCoder on GameGoData yields a model that outperforms matched baselines and is comparable to frontier models across gamedev benchmarks. All code, datasets, and models will be made publicly available. \noindent\url{https://github.com/Haoyue-Yang/GameGo}

\end{abstract}

\section{Introduction}
\label{sec:introduction} 
Recent advances in large language models (LLMs) enable the automatic generation of interactive frontend applications \cite{jiang2026opengame,li2026mm}. Building upon this trend, game development has been significant attentions, with recent benchmarks evaluating LLM agents across diverse core tasks, including game creation, debugging, and refinement \cite{chen2026gamexpert,chi2026gamedevbench}. While these efforts establish foundational evaluation frameworks, effective data synthesis for training game-generation agents remains underexplored. To bridge this gap, this work focuses on systematically scaling high-quality trajectories for single-turn game generation, where an LLM agent autonomously develops games and performs self-correction based on user queries. Under this setting, games generated directly from short and sparse queries often suffer from incomplete implementations, fragmented gameplay flows, and limited visual polish, as brief concepts leave critical execution decisions unresolved \cite{mu2024clarifygpt}. Conversely, enriching queries with structured gameplay and visual details guides agents toward markedly more complete and playable games.

These empirical observations highlight a core challenge: brief queries force coding agents to infer complex mechanics, progression, and visual representations without sufficient information, ultimately leading to suboptimal game artifacts. To resolve this issue, the paper presents \textbf{GameGo}, a requirements-generation pipeline grounded in industry game-development practices. Built upon real-world game seeds collected from diverse platforms, GameGo systematically enriches raw query through staged design decisions. Beginning with high-level ideas, it hierarchically structures the design, progressing from core gameplay loops and visual aesthetics to fine-grained specifications—including character roles, numerical balance, control schemes, camera behaviors, and state transitions. Asset planning further aligns these specifications with visual realizations, yielding a comprehensive Product Requirements Document (PRD) that provides coherent, actionable guidance for downstream code generation.


While comprehensive PRDs improve generated game completeness, over-specification constrains design exploration and overly long prompts degrade LLM instruction following. To address this trade-off, the paper further introduces task-adaptive query compression to retain only essential requirements that define core mechanics, playable loops, and visual presentation. By adapting the information budget to key gameplay dependencies, this approach eliminates redundant details while preserving execution freedom for coding agents. Compressed queries achieve a median 3.79-fold relation density gain over full PRDs while maintaining a concise length comparable to direct queries. Crucially, games generated from these high-density queries achieve an 85.7\% human preference over full PRD-generated games and an 86.6\% preference over direct-query baselines, demonstrating that explicitly modeling execution freedom yields superior game artifacts.

To evaluate the training utility of this framework, \textbf{GameGoData} is constructed, comprising 55,060 execution trajectories, and \textbf{GameGoBench} is established as a benchmark containing 124 game queries across diverse rendering formats and difficulty levels. Strict data decontaminatio is enforced by separating training and test seeds prior to task construction. Supervised fine-tuning of open-accessed models (e.g., Qwen3.5-27B and Qwen3.8-27B) on GameGoData yields \textbf{GameGoCoder} models, which consistently outperform matched base models across ArtifactsBench \cite{zhang2025artifactsbench}, Cookie-Bench \cite{yang2026cookie} and GameGoBench. These results demonstrate that training on trajectories distilled from structured game requirements enables small-parameter models to rival leading frontier models. In summary, our contributions are listed below:
\begin{itemize}[leftmargin=*]
    \item GameGo establishes a specification-synthesis framework mirroring real-world game development, applying dynamic constraint satisfaction to expand raw seeds and condense them into high-density, task-adaptive queries.
    \item Open-source models fine-tuned on GameGoData yield GameGoCoder, which consistently outperforms matched base models and achieves state-of-the-art results across public benchmarks and GameGoBench.
\end{itemize}

\begin{figure}[t]
    \centering
    \includegraphics[width=0.98\textwidth]{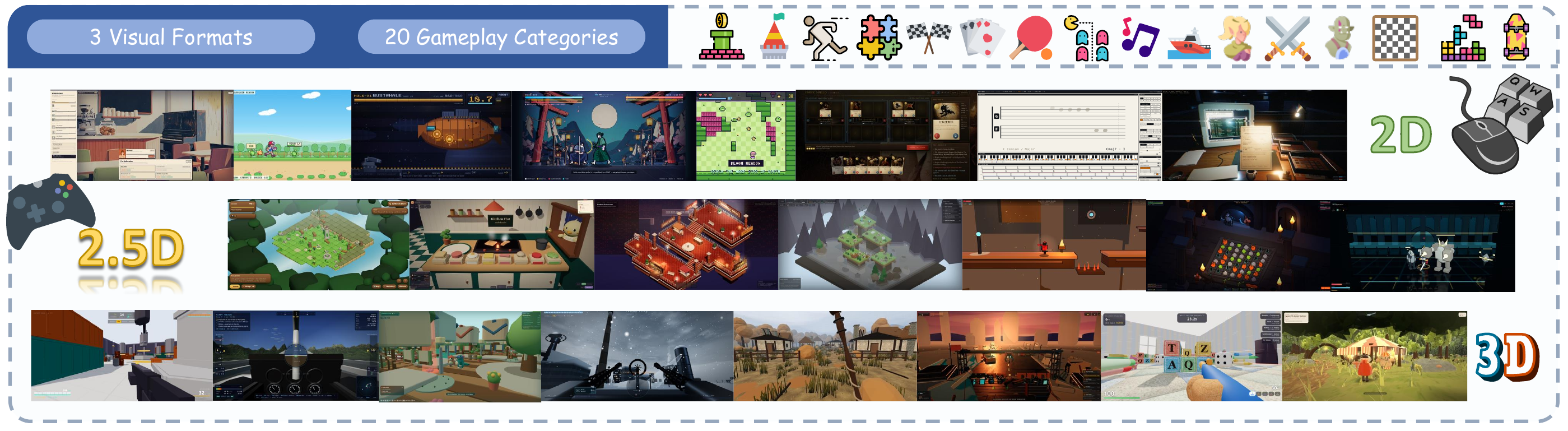}
    \vspace{-10pt}
    \caption{Game-development scope of GameGo. The tasks span 2D, 2.5D, and 3D formats and 20 gameplay categories, with representative runnable artifacts illustrating their diversity in appearance, interaction, and spatial structure.}
    \label{fig:gamego_coverage}
    \vspace{-10pt}
\end{figure}

\section{Related Work}
\label{sec:related_work}

\subsection{GameDev Benchmarks} 
Game-development benchmarks now evaluate more than whether generated code builds. GameDevBench \cite{chi2026gamedevbench} measures multimodal agents on complex changes to existing game projects, whereas WebGameBench \cite{zhang2026webgamebench} tests requirement-to-application delivery for browser-native games. GameXpert-Bench \cite{chen2026gamexpert} covers initial generation, bug repair, and multi-turn optimization, and GameCraft-Bench \cite{luo2026gamecraft} evaluates complete Godot games through replayed demonstrations and multimodal judgments. V-GameGym \cite{zhang2026v} studies visual game generation, while JAMER \cite{sun2026jamer} brings project-level evaluation to professional game engines. Related interactive-code benchmarks, including ArtifactsBench \cite{zhang2025artifactsbench} and Cookie-Bench \cite{yang2026cookie}, likewise use rendered behavior and autonomous interaction as evidence.


\subsection{GameDev Agents}
Existing gamedev agents improve the agent along several complementary axes. OpenGame \cite{jiang2026opengame} combines reusable game-development skills with a specialized coding model and execution-grounded optimization for end-to-end web games. ALIVE \cite{zhangbringing} uses automated play to curate training data and optimize frontend mini-game models, while GUI Agents for Continual Game Generation \cite{huang2026gui} couples coding with GUI playtesting to support iterative revision. CreativeGame \cite{ma2026creativegame} instead focuses on mechanic-aware generation. These methods strengthen the model, agent scaffold, or feedback loop once a task has been specified.

\vspace{-3pt}
\section{GameGo: Constructing Game-Development Tasks}
\label{sec:method}
\vspace{-3pt}

\begin{figure}[t]
    \centering
    \includegraphics[width=\textwidth]{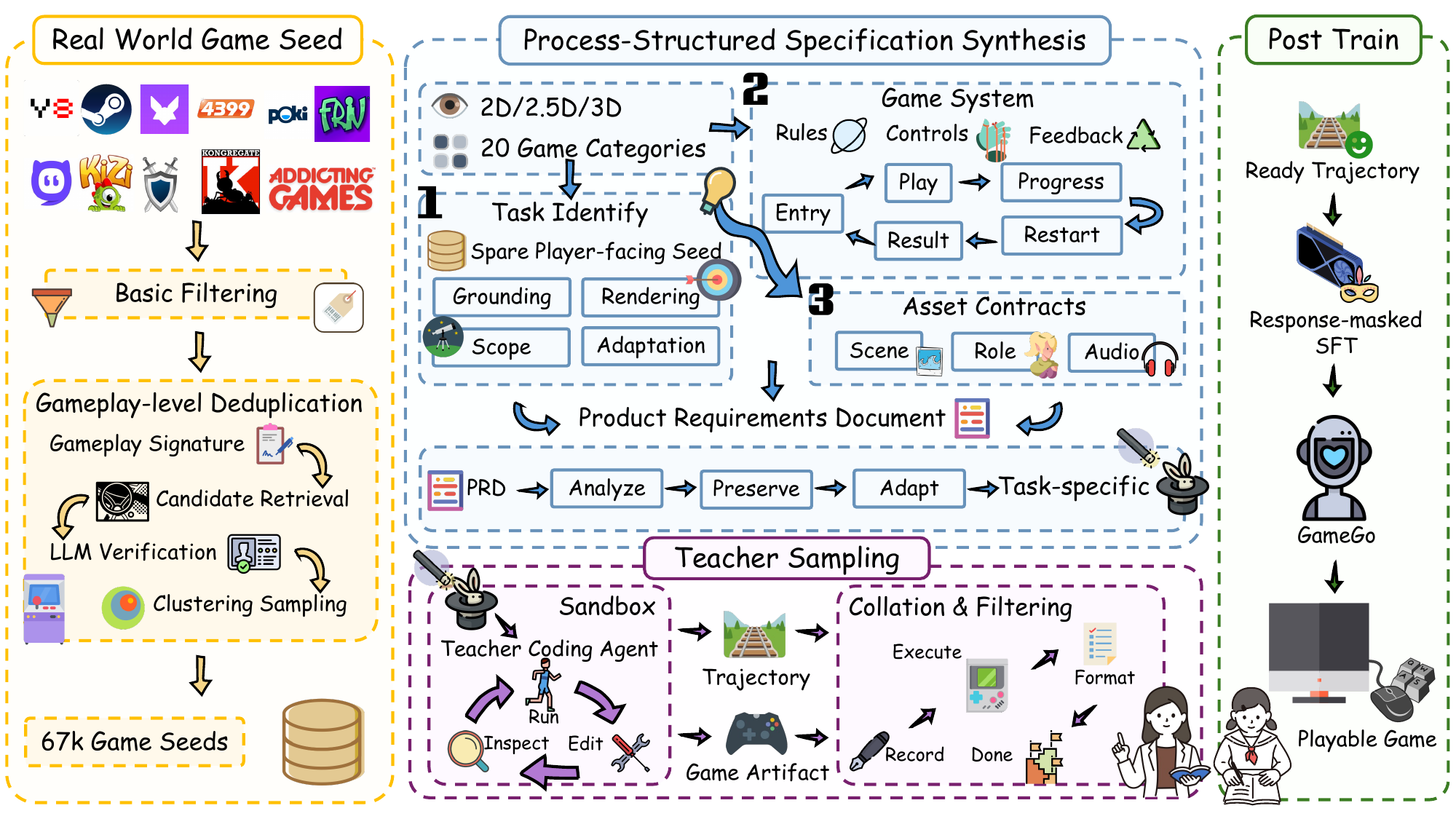}
    \vspace{-20pt}
    \caption{GameGo separates internal planning from execution instructions. Game seeds are expanded into PRDs that make gameplay and production dependencies explicit. Task-specific queries select the requirements delivered to a frontier coding agent, whose sandboxed development trajectories and game artifacts provide supervision for GameGoCoder.}
    \label{fig:gamego_pipeline}
    \vspace{-10pt}
\end{figure}

To construct game-development trajectories at scale, GameGo converts existing game catalogs into training tasks in four stages, as illustrated in Figure~\ref{fig:gamego_pipeline}. Game records are first collected from eighteen source entries, including descriptions, instructions, metadata, and available screenshots, and deduplicated at the gameplay level into a diverse seed pool. Staged planning then defines each seed's task scope, playable blueprint, and asset requirements, after which PRD compilation merges these outputs into a complete development specification. Task-adaptive query construction converts each PRD into a concise task description phrased as a natural user request, preserving the core mechanics while removing redundant implementation details and leaving routine choices open to the coding agent. Finally, a frontier model executes these queries in sandboxed environments, and each development trajectory is retained together with its final runnable artifact after execution and format checks, forming GameGoData.

\vspace{-3pt}
\subsection{Game Seed Collection}
\label{sec:seed_collection}
\vspace{-3pt}

To collect real-world game-development requirements, game records are collected from 18 source entries across three primary families: public Steam store pages and metadata\footnote{\url{https://store.steampowered.com/}}, 14 distinct source websites \footnote{Representative portals include \url{https://www.y8.com/} and \url{https://www.crazygames.com/}.}, and public WeChat mini-game sources\footnote{\url{https://www.17yoo.cn/}}. Each record retains its title, description, metadata, and available screenshots. All records are standardized into a fixed schema, promotional markup is stripped, and only those with clear gameplay rules, objectives, or interactions are retained, preserving category and genre fields as source metadata. Moreover, as these websites often contain repetitive or similar games, records are clustered by gameplay similarity, and a cluster-size-dependent quota is applied to limit redundancy while preserving small gameplay families. Specifically, records are considered redundant when they share the same core actions, state transitions, and objectives, differing only in presentation, parameters, or optional rules. For each record, a gameplay signature is extracted, candidate pairs are retrieved using lexical and semantic matching, and an LLM judge is employed to verify gameplay equivalence. The clustering and quota-based selection stage processes 70{,}000 records and retains 67{,}564 seeds. Appendix~\ref{app:gameplay_dedup} details the complete procedure and settings.

\vspace{-3pt}
\subsection{From Game Seeds to Executable Specifications}
\label{sec:specification_synthesis}
\vspace{-3pt}


For each retained game seed, a multi-stage compilation pipeline expands raw metadata into implementation-ready Product Requirements Documents. Drawing inspiration from real-world game development workflows and grounded in dynamic constraint satisfaction \cite{mittal1990dynamic}, specification synthesis operates via three planning stages and canonical compilation, coordinating activity constraints that determine required design dimensions with compatibility constraints that govern their integration.


\textbf{Task definition.}
Raw seed descriptions leave design boundaries ambiguous, causing planning drift toward generic conventions. The first stage applies activity constraints to convert raw seeds into concise task definitions, extracting core mechanics while stripping proprietary identities and branding. Query-associated screenshots or text descriptions are parsed to anchor visual evidence. It fixes the gameplay category, rendering format, and primary play phases, dynamically activating downstream planning routes without over-constraining implementation choices.

\begin{wrapfigure}{r}{0.3\columnwidth}
    \centering
    \includegraphics[width=\linewidth]{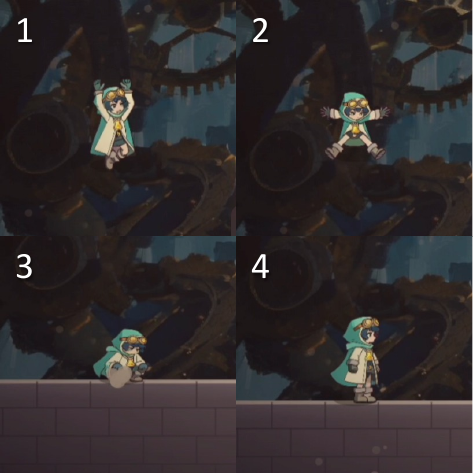}
    \vspace{-20pt}
    \caption{Synthesized character sprite sheet for a side-scrolling platformer.}
    \label{fig:sprite_animation_example}
    \vspace{-15pt}
\end{wrapfigure}


\textbf{Playable blueprinting.}
Mechanics alone do not specify how a game progresses or recovers from failure. Building on active task definitions, the second stage enforces behavioral compatibility constraints across the full gameplay flow. It specifies camera behavior, spatial relations, controls, failure conditions, and state transitions from entry through restart. Domain-specific guidance conditions this planning, such as mandating multi-frame character sprite sheets for side-scrolling actions to guide keyframe transitions, as illustrated in Figure~\ref{fig:sprite_animation_example}. A blueprint is complete when its state transitions form a closed, compatible playable loop.


\textbf{Asset grounding.}
Behavioral specifications leave unresolved how game entities, scenes, and feedback are realized. Once gameplay structure is fixed, the third stage grounds resource requirements under structural compatibility constraints. It maps each resource to a stable project path, production source, dimension budget, and execution method across procedural geometry, textures, audio, and shaders. Foreground interactable entities are strictly separated from background elements, establishing deterministic asset contracts before code generation.


\textbf{PRD compilation.}
The final step synthesizes stage outputs into a single canonical document via field-based integration. Task definitions supply scope, blueprinting provides state transitions, and asset grounding supplies visual contracts. The compiler selects authoritative fields and eliminates redundant descriptors while preserving cross-stage dependencies, resolving potential constraint conflicts to provide a comprehensive planning representation for task-adaptive query construction.

\subsection{Task-Adaptive Query Construction}
\label{sec:query_construction}

While full-length PRDs guarantee structural coverage, their extensive length overburdens instruction following and restricts design exploration for downstream coding agents. To resolve this limitation, a task-adaptive construction method compiles concise execution queries \(q\) from comprehensive requirements \(p\) under an information budget calibrated to structural complexity. For a seed \(s\) containing \(n_s\) reference gameplay clauses in \(p\), let \(L(x)\) denote the token count and \(K_s(x)\) represent the number of clauses fully captured in \(x\). The token retention ratio is defined as \(r_s = L(q)/L(p)\) and relation retention as \(R_s = K_s(q)/n_s\). The relation density \(\rho_s(x) = K_s(x)/L(x)\) quantifies captured clauses per token, yielding a density gain \(G_s = \rho_s(q)/\rho_s(p) = R_s/r_s\), where \(G_s > 1\) reflects elevated information density.

This formulation prunes verbose implementation noise while preserving gameplay-defining constraints under the assigned budget. It retains core parameters, including identity, spatial relations, control schemes, and failure modes, while removing redundant descriptions and exhaustive asset lists. Each query includes an audit log tracking preserved constraints, open design choices, and complexity rationales. Independent LLM agents validate query fidelity and execution freedom, triggering iterative refinement for non-compliant outputs. Optional stylistic directions are assigned to a sampled task subset, as detailed in Appendix~\ref{app:query_construction}. Beyond standard prompt compression, this approach explicitly models creative design freedom within the task representation.

\section{GameGoData and GameGoBench}
\label{sec:training}

Following prior work, a frontier model is employed to execute each task query, generating the final game artifact while capturing the full development trajectory as process supervision. These collected trajectories form GameGoData, which is used to train GameGoCoder. Additionally, GameGoBench is constructed, comprising 124 diverse game-development queries for comprehensive evaluation.

\subsection{Trajectory Synthesis}
\label{sec:sandbox}

\begin{figure}[t]
    \centering
    \includegraphics[width=0.75\textwidth]{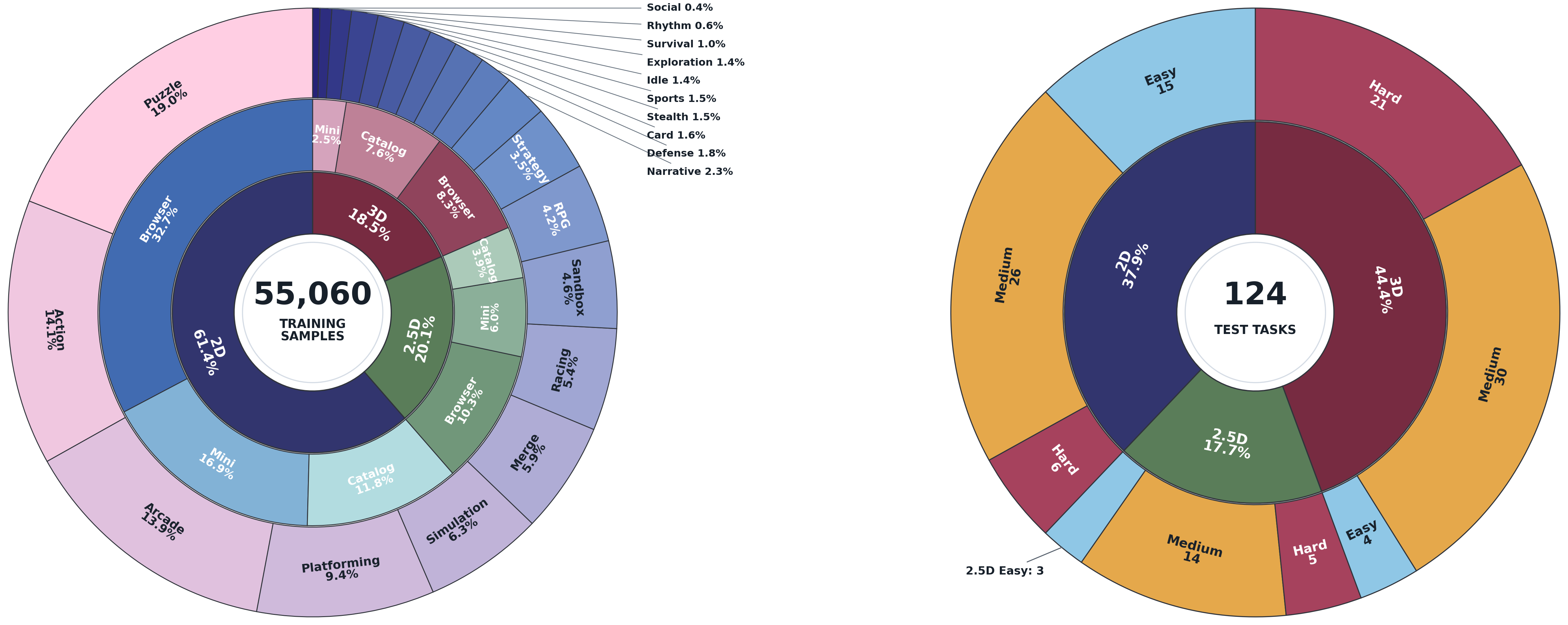}
    \vspace{-3mm}
    \caption{Left: GameGoData composition by visual format, source group, and gameplay category. Right: GameGoBench composition, including visual format and format-conditioned difficulty.}
    \label{fig:train&test_data}
    \vspace{-10pt}
\end{figure}

To enable complex interaction execution, an agent scaffold adapted from the Code Arena framework\footnote{\url{https://arena.ai/leaderboard/code/webdev}} is deployed inside isolated browser-game sandboxes. Configured on a standardized React template, this scaffold equips the agent with a general-purpose programming toolset tailored for web-game development, including file-system editing, project compilation via \texttt{build\_project}, and interactive shell execution. During execution, GameGoData systematically logs full agent interaction trajectories, including text messages, tool calls, tool responses, and final artifacts. To eliminate non-functional execution trajectories that yield invalid supervision, a multi-stage filtering protocol is applied to retain only executable projects with complete codebases, valid parsed structures, and verified runtime success. To this end, from 67{,}564 submitted task-specific queries, 67{,}487 initial trajectories are recorded. Among these, 65{,}082 complete the primary execution phase successfully, achieving a 96.3\% completion rate. Subsequent artifact integrity and formatting verification retain 55{,}060 examples, corresponding to an overall yield of 81.5\% for GameGoData. The final dataset contains 61.4\% 2D, 20.1\% 2.5D, and 18.5\% 3D tasks, with complete compositional distributions summarized in Figure~\ref{fig:train&test_data}. Appendix~\ref{app:react_quality_checks} details the admission rules and post-build truncation.



\subsection{GameGoBench}
\textbf{Task Source.}
In parallel with GameGoData, GameGoBench is established as a held-out benchmark for rigorous evaluation of game-development agents. It comprises 124 diverse tasks spanning 2D (47), 2.5D (22), and 3D (55) rendering formats. Tasks within each visual format are stratified into Easy, Medium, and Hard difficulty levels based on their implementation complexity, as detailed in Figure~\ref{fig:train&test_data}. To provide a holistic measurement, each task in the benchmark is paired with a multi-dimensional verification protocol to assess the generated game artifacts.

\textbf{Artifact Evaluation.}
To rigorously evaluate the synthesized games, we construct an automated dynamic verification environment, as static code analysis fails to capture interactive runtime behaviors. Specifically, each game artifact is executed in a headless browser, where an interactive multimodal LLM agent performs step-driven actions, advancing the game clock only upon agent input. Another independent evaluation agent then scores the recorded interaction trace based on three key aspects. First, Execution Rate measures the proportion of generated repositories that successfully render a page in the browser. Second, Requirements measures functional fidelity through a four-level hierarchy covering core concepts, input responsiveness, game loops, and query specifications. Finally, Quality assesses visual aesthetics across subject art, scene composition, and UI integration, using chronologically sampled frames to capture representative gameplay rather than cherry-picked states.



\section{Experiments}
\label{sec:experiments}

\subsection{Implementation Details}
\label{sec:implementation_details}
To evaluate the quality of the constructed game-development trajectory, we perform SFT on Qwen3.5-27B \cite{qwen2026qwen35} and Qwen3.8-27B \cite{qwen2026qwen38max} as the base models with GameGoData. Specifically, we train all parameters of the language model for one epoch on $55,060$ length-filtered trajectories in BF16 precision, while keeping the vision encoder and aligner frozen. To focus optimization on task execution, the response-masked objective computes loss exclusively over assistant thoughts, responses, and tool calls, treating system prompts, user queries, and tool responses as context. As GameGoData contains no image inputs, the vision encoder is not used. Training runs with sequence packing at an effective global batch size of 8, a 256K context window, and a peak learning rate of $1 \times 10^{-5}$. At inference, models execute within the Code Arena scaffold in isolated sandboxes, generating one rollout per query. Appendix~\ref{app:evaluation_design} and \ref{app:experimental_details} detail the two-stage evaluation protocol and extended training setups.

\subsection{Evaluation Settings}
GameGoCoder is evaluated on two public benchmarks: the game portions of ArtifactsBench \cite{zhang2025artifactsbench} and Cookie-Bench \cite{yang2026cookie} which are denoted as ArtifactsBench-G and Cookie-Bench-G. Their official queries and evaluation pipelines are retained, with Overall reported on 413 ArtifactsBench-G tasks and Functionality, Aesthetics, and Overall on 156 Cookie-Bench-G tasks. For all three benchmarks, generated React repositories are evaluated with Claude Sonnet 4.6 powering the respective verifier pipelines. Appendix~\ref{app:verifier_scope} details the verifier sensitivity and failure handling procedures.

For performance comparison, we choose six large-scale models includes: Claude Opus 5 \cite{anthropic2026opus5}, GPT-5.6 \cite{openai2026gpt56}, DeepSeek-V4-Pro \cite{deepseek2026v4}, GLM-5.3 \cite{zai2026glm53}, Qwen3.8-Max \cite{qwen2026qwen38max}, and Kimi-K3 \cite{kimi2026k3} as the leading performnace. Besides large-scale models, we also compare with Qwen3.5-27B \cite{qwen2026qwen35} and Qwen3.8-27B \cite{qwen2026qwen38max} with corresponding GameGoCoder checkpoints. Within each benchmark, all models use the same Code Arena scaffold, sandbox environment, and inference budget. Additional training and evaluation details are given in Appendix~\ref{app:experimental_details}.

\begin{table}[t]
\caption{Main results on game development benchmarks. Req. and Quality denote requirement satisfaction and game quality scored by the Claude Sonnet 4.6 verifier.}
\label{tab:main_results}
\centering
\footnotesize
\setlength{\tabcolsep}{2.0pt}
\renewcommand{\arraystretch}{1.05}
\begin{tabular*}{\textwidth}{@{\extracolsep{\fill}}lcccccccc@{}}
\toprule
& \multicolumn{1}{c}{ArtifactsBench-G}
& \multicolumn{3}{c}{CookieBench-G}
& \multicolumn{4}{c}{GameGoBench} \\
\cmidrule(lr){2-2}\cmidrule(lr){3-5}\cmidrule(lr){6-9}
Model & Overall
& Func. & Aesth.  & Overall 
& Exec. & Req.  & Quality & Overall \\
\midrule
\multicolumn{9}{l}{\emph{Frontier models}} \\
Claude Opus 5 & 72.76 & 94.63 & 86.25 & 83.68 & 98.32 & 68.18 & 55.50 & 61.84 \\
GPT-5.6 & 63.92 & 86.38 & 85.13 & 75.36 & 100.00 & 59.57 & 45.99 & 52.78 \\
DeepSeek-V4-Pro & 62.77 & 91.50 & 80.88 & 77.07 & 99.19 & 56.45 & 42.60 & 49.52 \\
GLM-5.3 & 68.63 & 91.50 & 85.13 & 79.99 & 99.18 & 68.85 & 50.77 & 59.81 \\
Qwen3.8-Max & 66.92 & 93.88 & 90.38 & 85.96 & 100.00 & 64.52 & 52.19 & 58.35 \\
Kimi-K3 & 61.95 & 93.75 & 86.63 & 83.63 & 100.00 & 60.94 & 46.69 & 53.82 \\
\midrule
\multicolumn{9}{l}{\emph{Baselines and GameGoCoder}} \\
\shortstack[l]{Qwen3.5-27B} & 46.31 & 64.25 & 56.63 & 41.35 & 91.80 & 36.89 & 31.12 & 34.00 \\
\textbf{GameGoCoder 3.5} & 57.34 & 65.50 & 69.88 & 55.79 & 98.29 & 57.99 & 45.72 & 51.85 \\
\shortstack[l]{Qwen3.8-27B} & 59.87 & 87.38 & 85.38 & 77.07 & 100.00 & 62.81 & 42.25 & 52.53 \\
\textbf{GameGoCoder 3.8} & 61.94 & 90.63 & 85.75 & 80.68 & 100.00 & 66.94 & 43.64 & 55.29 \\

\bottomrule
\end{tabular*}
\vspace{-5pt}
\end{table}

\subsection{Main Results}
Table~\ref{tab:main_results} compares GameGoCoder with frontier models and matched base models on the two public benchmarks and GameGoBench. GameGoCoder improves over both matched base models across the public benchmarks and the held-out GameGoBench, with the largest gains obtained from the Qwen3.5-27B backbone. The stronger Qwen3.8-27B backbone also benefits consistently from GameGo post-training. Our qualitative showcase in Figures~\ref{fig:showcase_part1} and~\ref{fig:showcase_part2} compares ten models across twelve tasks under identical queries, illustrating differences in scene composition, visual detail, interface presentation, and implementation completeness.

\begin{wrapfigure}{r}{0.48\columnwidth}
    \centering
    \includegraphics[width=\linewidth]{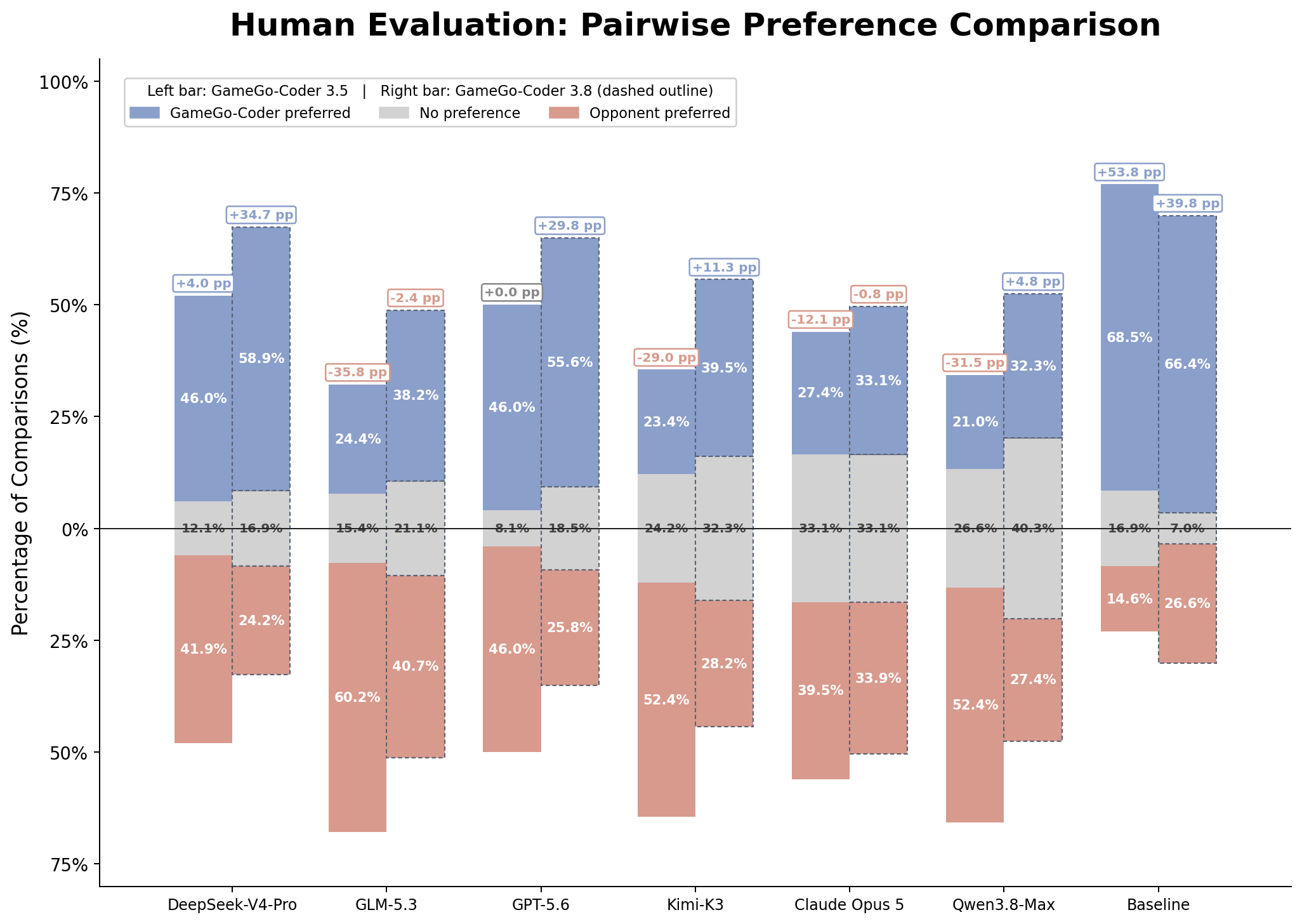}
    \vspace{-20pt}
    \caption{Human pairwise preference evaluation against the models on the horizontal axis. Each group shows GameGoCoder 3.5 on the left and 3.8 as the dashed right bar.}
    \label{fig:pairwise_preference}
    \vspace{-10pt}
\end{wrapfigure}

\begin{figure*}[t]
    \centering
    \includegraphics[width=0.98\textwidth]{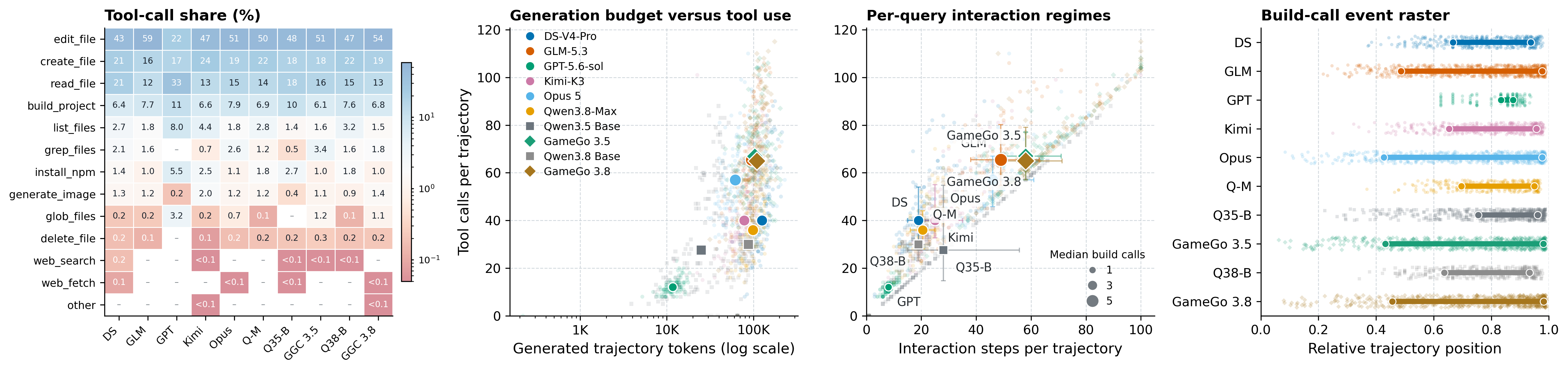}
    \vspace{-10pt}
    \caption{Trajectory analysis over 124 queries and models. From left to right: tool-use shares; generated tokens versus tool calls; interaction regimes; and \texttt{build\_project} positions.}
    \label{fig:main_trajectory_analysis}
    \vspace{-10pt}
\end{figure*}

Given that LLM-based evaluators remain susceptible to score hacking, we further incorporate human expert oversight. Following Code Arena, six hired independent expert evaluators provide blinded pairwise judgments on GameGoBench, with the protocol and agreement results detailed in Appendix~\ref{app:human_evaluation}. From the 124 tasks, reviewers compare available playable artifact pairs from GameGoCoder and each comparison model, with model identities hidden, then select the preferred artifact or a tie. As shown in Figure~\ref{fig:pairwise_preference}, human preferences broadly track the automated results: both GameGoCoder variants are clearly favored over their matched base models, while GameGoCoder 3.8 is strongly favored over several frontier models and remains competitive with the strongest ones.

Beyond benchmark scores, model trajectories are analyzed on GameGoBench across 124 shared queries and ten models. Figure~\ref{fig:main_trajectory_analysis} summarizes tool-call shares, output tokens versus tool use, per-query interaction regimes, and the relative positions of \texttt{build\_project} calls. Tool use is dominated by file editing, creation, and reading, while the scatter plots show that models with similar outcomes can adopt different interaction budgets. Compared with its matched base models, GameGoCoder spends more interaction on implementation and verification. Also, frontier models generally reach comparable outcomes with shorter trajectories, and the raster of \texttt{build\_project} calls further reveals differences in when verification occurs. These paired diagnostics complement the aggregate benchmark scores and are not included in them.

\vspace{-3pt}
\subsection{Ablation Studies}
\vspace{-3pt}

\subsubsection{Query Construction Ablations.}
\label{sec:ablation_analysis}

\vspace{-3pt}
To demonstrate the effectiveness of task-specific queries, we compare them against direct queries and full-length PRD methods. 
However, synthesizing full-scale trajectories incurs substantial API costs and long rollout latency.
To manage compute budget while maintaining rigorous evaluation, we restrict format comparisons to the 124 tasks of GameGoBench with the following three aspects.

\textbf{Artifact quality.}
We first ablate on artifact quality generated by each query format. Six independent expert evaluators conduct pairwise comparisons across fidelity and playability. Table~\ref{tab:pipeline_ablation} aggregates win, tie, and loss judgments across artifact pairs from the perspective of the first representation, where preference is defined as \(W/(W+L)\), excluding ties. Planning enhances artifact preference, and selective compression provides further gains. This evaluation measures the combined query-construction procedure, with downstream SFT relations and cost rationales detailed in Appendix~\ref{app:evaluation_design}.


\textbf{Compression Analysis.}
\label{sec:static_analysis_results}
We then analyze the compression strategy across 124 GameGoBench tasks, where direct and task-specific queries show median lengths of 227.5 and 235 tokens, respectively. Median retained length relative to full PRDs is 2.26\%, yielding \(44.2\times\) compression, or \(21.6\times\) excluding asset appendices. An AI-assisted audit of 927 PRD clauses identifies 12.1\% as fully expressed, 62.2\% as partially expressed, and 25.7\% as absent. Under the relation density formulation \(G_s = R_s / r_s\), counting complete clauses indicates a median density gain \(G_s\) of \(3.79\times\), or \(1.78\times\) relative to PRD bodies. Paired with human preferences of 85.7\% over full PRDs and 86.6\% over direct queries, task-specific queries substantially reduce text volume while generating superior game artifacts. Comprehensive clause audit parameters are provided in Appendix~\ref{app:static_specification_analysis}.

\begin{table}[t]
\centering
\caption{Human pairwise evaluation of three task representations on 124 GameGoBench tasks. Win, tie, and loss cells give pooled counts with percentages of all reviewer judgments in parentheses.}
\label{tab:pipeline_ablation}
\small
\setlength{\tabcolsep}{5pt}
\begin{tabular}{@{}lcccc@{}}
\toprule
Pairwise comparison & Win & Tie & Loss & Preference \\
\midrule
full PRD vs. direct & 369 (49.6) & 209 (28.1) & 166 (22.3) & 0.690 \\
task-specific vs. full PRD & 487 (65.5) & 176 (23.7) & 81 (10.9) & 0.857 \\
task-specific vs. direct & 606 (81.5) & 44 (5.9) & 94 (12.6) & 0.866 \\
\bottomrule
\end{tabular}
\vspace{-10pt}
\end{table}

\begin{figure}[t]
    \centering
    \includegraphics[width=0.98\textwidth]{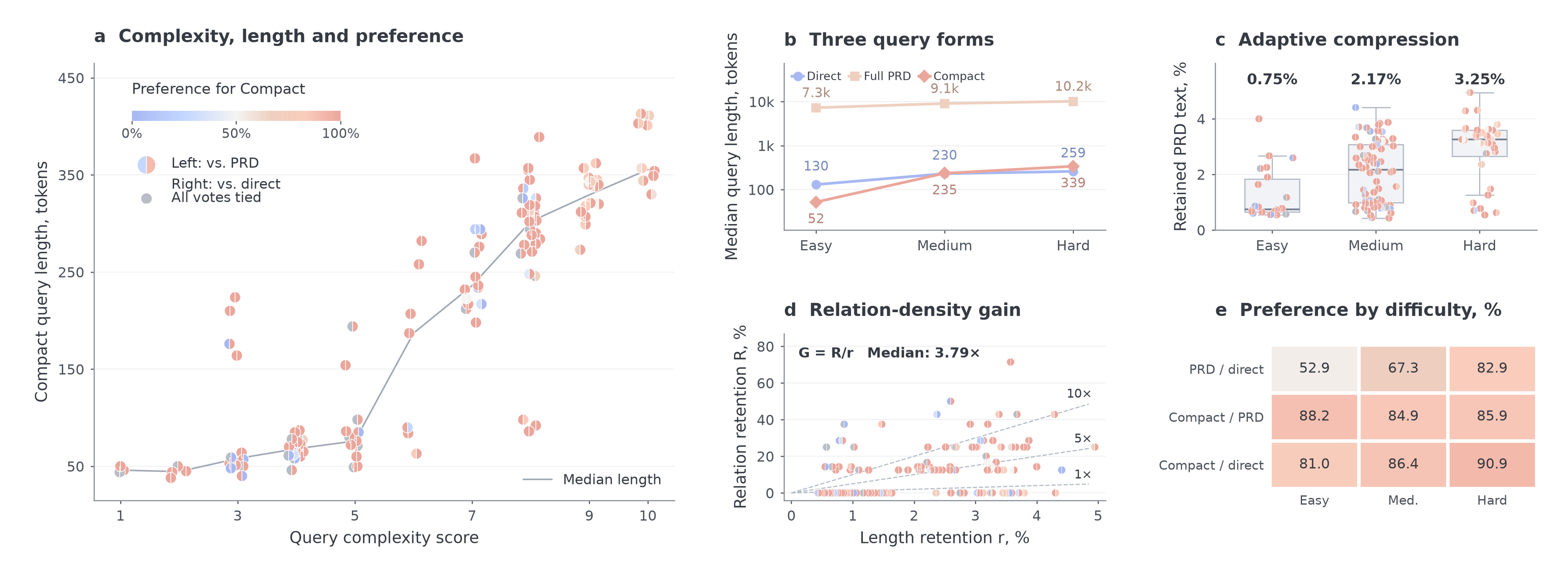}
    \vspace{-15pt}
    \caption{Task-adaptive compression and artifact preference across 124 games. Compact denotes task-specific queries. Marker halves depict Compact preference over full PRDs and direct queries, with gray indicating tied votes.}
    \label{fig:task_query_overview}
    \vspace{-10pt}
\end{figure}

\textbf{Complexity-Adaptive Allocation.}
Figure~\ref{fig:task_query_overview} provides task-level and difficulty-stratified comparisons. Task-specific query lengths scale dynamically with complexity scores (panel a), expanding from 52 tokens (Easy) to 235 (Medium) and 339 (Hard), with retained PRD fractions increasing from 0.75\% to 3.25\% (panels b--c). Compared to direct queries, task-specific queries remain shorter on Easy tasks (52 vs. 130 tokens) but expand on Hard tasks (339 vs. 259), confirming demand-driven capacity allocation over uniform length minimization. Consequently, artifact preferences consistently favor task-specific queries across Easy, Medium, and Hard tiers, achieving up to 88.2\% win rates against full PRDs and 90.9\% against direct queries (panel e). These findings confirm that complexity-adaptive query detail yields superior artifact quality across all difficulty groups.

\subsubsection{Model Ablations.}
Since our primary evaluations focus on post-trained models, we conduct a training ablation on a raw base model to examine whether GameGoData enables effective learning without prior instruction tuning. Specifically, we fine-tune Qwen3.5-35B-A3B-MoE-Base, an open-weight base model of comparable size without prior SFT, on GameGoData for six epochs. This evaluate whether the data also supports learning before general-purpose instruction tuning. All checkpoints share the evaluation scaffold, sandbox, and interaction budget. We compare trajectory statistics for Base, the official post-trained Qwen3.5-35B-A3B-MoE, and Epochs~1 to 6 on 124 aligned GameGoBench tasks, complemented by blinded human comparisons of playable artifacts. 

\begin{figure*}[t]
    \centering
    \includegraphics[width=0.98\textwidth]{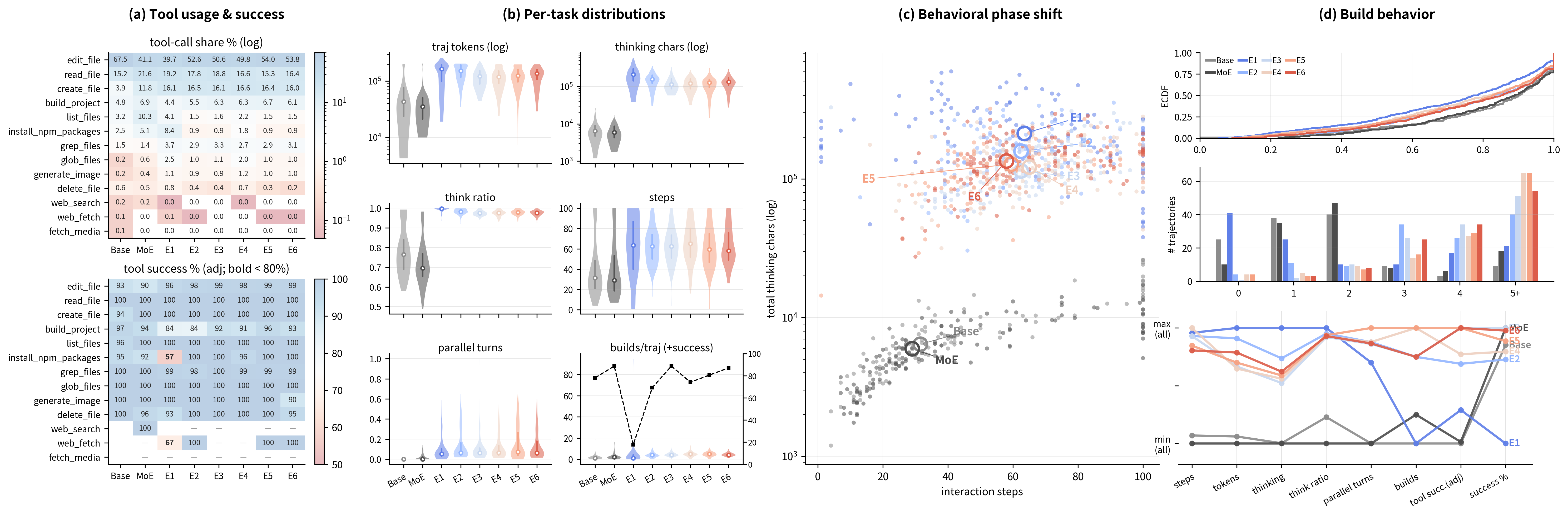}
    \vspace{-10pt}
    \caption{Training dynamics of the base model on 124 aligned tasks. Base and MoE denote the Qwen3.5-35B-A3B-Base and Qwen3.5-35B-A3B, respectively. E1 to E6 denote models saved after successive GameGo training epochs. From left to right: tool usage and success, per-task trajectory distributions, behavioral phase shift, and build behavior.}
    \label{fig:long_training_dynamics}
    \vspace{-10pt}
\end{figure*}

We first analyze the training phases in Figure~\ref{fig:long_training_dynamics}. Initially, trajectories lengthen through additional reasoning tokens while primary success drops, suggesting that longer deliberation does not yet translate into working implementations. Later, the model invokes builds earlier, uses tools more reliably, and performs more cycles of building, verification, and repair. The first panel also shows a shift from directory listing toward file creation. By the end, every evaluated trajectory contains a build and primary success recovers non-monotonically. These dynamics suggest that GameGo supervision first lengthens reasoning, then improves its use of execution feedback. Performance therefore improves unevenly over training rather than increasing with reasoning length alone.

\begin{wraptable}{r}{0.48\columnwidth}
      \vspace{-20pt}
      \centering
      \scriptsize
      \setlength{\tabcolsep}{3pt}
      \caption{Blinded human preference for GameGo-trained MoE checkpoints. $N$ includes pairs with playable artifacts.}
      \label{tab:moe_pairwise_ablation}
      \resizebox{\linewidth}{!}{%
      \begin{tabular}{@{}lccc@{}}
      \toprule
      Comparison & $N$ & W/T/L & Preference \\
      \midrule
      E2 ckpt vs. Base & 110 & 437/147/76 & 0.852 \\
      E2 ckpt vs. MoE & 108 & 355/113/180 & 0.664 \\
      \midrule
      E6 ckpt vs. MoE & 123 & 453/177/108 & 0.807 \\
      \bottomrule
      \end{tabular}%
      }
      \vspace{-8pt}
  \end{wraptable}

Six independent expert evaluators further conduct blinded comparisons on executable game pairs.  The results are shown in Table~\ref{tab:moe_pairwise_ablation}, while W/T/L pools judgments from the former model perspective, and the preference also excludes ties. The results show that Epoch~2 is preferred to Base with a score of 0.852 and to the official post-trained checkpoint with 0.664, supporting improved artifact quality over base and post-trained models among playable pairs. Furthermore, Epoch~6 provides a longer-training reference, scoring 0.807 against the official model. We further detail the protocol and reviewer agreement in the Appendix~\ref{app:human_evaluation}.

\vspace{-8pt}
\section{Conclusion}
\vspace{-8pt}
 
Generating games from raw seeds balances over-specified restrictions against under-specified instruction drift. GameGo addresses this trade-off by combining seed-to-PRD expansion with dynamic constraint compression, separating design planning from execution instructions. High-density compressed queries achieve superior human preference over both full-PRD and direct-query baselines. Applying this pipeline produces GameGoData with 55,060 execution trajectories across 2D, 2.5D, and 3D formats alongside a strictly disjoint GameGoBench. Training GameGoCoder on GameGoData yields models that consistently outperform matched baselines and rival frontier models across game-development benchmarks. These results establish structured specification synthesis paired with task-adaptive compression as a practical paradigm for training game-development agents.

\section*{AI Use Statement}
In this work, we used generative AI tools to clean and reformat dataset artifacts during training corpus construction and edit text to improve readability. Generative AI tools were not used for developing theoretical models, formulating mathematical claims, assisting in proof writing, proposing hypotheses, designing experimental methodology, or interpreting results, and all remaining required disclosure tasks are not applicable. All AI-assisted output was thoroughly reviewed and validated by the authors, including strict auditing of dataset integrity. We take full responsibility for the final content of this work, including text and artifacts produced with the aid of generative AI.

\section*{Ethics Statement}

To the best of our knowledge, this work does not raise additional ethical concerns beyond those addressed by the ICLR Code of Ethics. The study employs independent expert evaluators to compare generated game artifacts under the protocols described in the paper. The data and resources used in this work were obtained and used in accordance with their applicable licenses and terms of use. We have considered potential risks related to privacy, security, bias, fairness, misuse, and societal impact, and we are not aware of any unresolved issue that would prevent the responsible dissemination of the results. All authors have read and agree to adhere to the ICLR Code of Ethics.

\section*{Reproducibility Statement}
To ensure the reproducibility of this work, comprehensive implementation details, dataset processing protocols, and evaluation environments are provided across the main text and appendices. Complete algorithmic procedures for specification expansion and dynamic constraint compression are detailed in the methodology section. Full details regarding training corpus curation, de-duplication filtering, and quality verification for GameGoData are documented in the dataset appendix. Exact training hyperparameters, prompt templates, model configurations, and interaction budgets are supplied in the training and evaluation appendix.

\bibliography{iclr2027_conference}

@article{mu2024clarifygpt,
  title={Clarifygpt: A framework for enhancing llm-based code generation via requirements clarification},
  author={Mu, Fangwen and Shi, Lin and Wang, Song and Yu, Zhuohao and Zhang, Binquan and Wang, ChenXue and Liu, Shichao and Wang, Qing},
  journal={Proceedings of the ACM on Software Engineering},
  volume={1},
  number={FSE},
  pages={2332--2354},
  year={2024},
  publisher={ACM New York, NY, USA}
}

@article{li2026mm,
  title={MM-WebAgent: A Hierarchical Multimodal Web Agent for Webpage Generation},
  author={Li, Yan and Zeng, Zezi and Yang, Yifan and Yang, Yuqing and Liao, Ning and Guo, Weiwei and Qiu, Lili and Cheng, Mingxi and Dai, Qi and Wang, Zhendong and others},
  journal={arXiv preprint arXiv:2604.15309},
  year={2026}
}

@article{jiang2026opengame,
  title={Opengame: Open agentic coding for games},
  author={Jiang, Yilei and Hu, Jinyuan and Xiao, Qianyin and Zheng, Yaozhi and Ma, Ruize and Feng, Kaituo and Han, Jiaming and Peng, Tianshuo and Fan, Kaixuan and Zhang, Manyuan and others},
  journal={arXiv preprint arXiv:2604.18394},
  year={2026}
}

@inproceedings{zhangbringing,
  title={Bringing Code ALIVE: Optimizing Interactive Frontend Mini-Games via Automated Play and Reinforcement Learning at Scale},
  author={Zhang, Jiajun and Jing, Yuheng and Cui, Zeyu and Zheng, Hao and Chen, Wentao and Li, Kaixin and Yang, Jiaxi and Xie, Tianbao and Ma, Zeyao and Bai, Tianyi and others},
  booktitle={Forty-third International Conference on Machine Learning}
}

@article{chen2026gamexpert,
  title={GameXpert-Bench: How Far Are Coding Agents from Expert Game Development?},
  author={Chen, Kun and Hong, Haorong and Gao, Peizhong and Lin, Jianfeng and Luo, Tongxu and Xie, Yuxuan and Liu, Chenxu and He, Jieling and Liu, Zhongyuan and Zeng, Zeno},
  journal={arXiv preprint arXiv:2608.21833},
  year={2026}
}

@article{zhang2025artifactsbench,
  title={Artifactsbench: Bridging the visual-interactive gap in llm code generation evaluation},
  author={Zhang, Chenchen and Li, Yuhang and Xu, Can and Liu, Jiaheng and Liu, Ao and Zhou, Changzhi and Deng, Ken and Wu, Dengpeng and Huang, Guanhua and Li, Kejiao and others},
  journal={arXiv preprint arXiv:2507.04952},
  year={2025}
}

@article{yang2026cookie,
  title={Cookie-Bench: Continuous On-screen Key Interaction Evaluation for Web Generation},
  author={Yang, Haoyue and Shen, Zhangxiao and Ding, Fan and Lou, Hangting and Kou, Yifeng and Yu, Haoqing and Li, Jingyao and Wu, Zhengfan and Bao, Siqi and Liu, Jing and others},
  journal={arXiv preprint arXiv:2605.30000},
  year={2026}
}

@inproceedings{zhang2026v,
  title={V-gamegym: Visual game generation for code large language models},
  author={Zhang, Wei and Yang, Jian and Tao, Renshuai and Chai, Linzheng and Guo, Shuyue and Wu, Jiajun and Chen, Xiaoming and Cui, Ganqu and Ding, Ning and Xu, Xander and others},
  booktitle={Findings of the Association for Computational Linguistics: ACL 2026},
  pages={5613--5641},
  year={2026}
}

@article{zhang2026webgamebench,
  title={WebGameBench: Requirement-to-Application Evaluation for Coding Agents via Browser-Native Games},
  author={Zhang, Wenyu and You, Guoliang and Zhao, Haotian and Zhu, Tianshu and Wang, Haoran and Tang, Xiaoxuan and Dai, Mingyang and Gu, Jingnan and Dong, Daxiang and Wu, Jianmin and others},
  journal={arXiv preprint arXiv:2605.17637},
  year={2026}
}

@article{huang2026gui,
  title={GUI Agents for Continual Game Generation},
  author={Huang, Yixu and Li, Bo and Li, Na and Wang, Zhe and Chen, Kaijie and Ge, Haonan and Si, Qingyi and Shen, Yuanzhe and Yang, Ruihan and Wang, Guangjing and others},
  journal={arXiv preprint arXiv:2605.28258},
  year={2026}
}

@article{ma2026creativegame,
  title={CreativeGame: Toward Mechanic-Aware Creative Game Generation},
  author={Ma, Hongnan and Wang, Han and Wang, Shenglin and Yin, Tieyue and Shi, Yiwei and Huang, Yucong and Zou, Yingtian and Wen, Muning and Yang, Mengyue},
  journal={arXiv preprint arXiv:2604.19926},
  year={2026}
}

@article{sun2026jamer,
  title={JAMER: Project-Level Code Framework Dataset and Benchmark on Professional Game Engines},
  author={Sun, Jianwen and Li, Chuanhao and Li, Zizhen and Feng, Yukang and Zhang, Fanrui and Huang, Yifei and Dai, Yu and Zhang, Kaipeng},
  journal={arXiv preprint arXiv:2606.19830},
  year={2026}
}

@inproceedings{mittal1990dynamic,
  title={Dynamic constraint satisfaction},
  author={Mittal, Sanjay and Falkenhainer, Brian},
  booktitle={Proceedings eighth national conference on artificial intelligence},
  pages={25--32},
  year={1990}
}

@article{chi2026gamedevbench,
  title={Gamedevbench: Evaluating agentic capabilities through game development},
  author={Chi, Wayne and Fang, Yixiong and Yayavaram, Arnav and Yayavaram, Siddharth and Karten, Seth and Wei, Qiuhong Anna and Chen, Runkun and Wang, Alexander and Chen, Valerie and Talwalkar, Ameet and others},
  journal={arXiv preprint arXiv:2602.11103},
  year={2026}
}

@article{luo2026gamecraft,
  title={GameCraft-Bench: Can Agents Build Playable Games End-to-End in a Real Game Engine?},
  author={Luo, Tongxu and Wang, Rongsheng and Bi, Jiaxi and Xu, Chenming and Tang, Zhengyang and Chen, Jianlong and Liang, Juhao and Ji, Ke and Guo, Shuqi and Du, Yuhao and others},
  journal={arXiv preprint arXiv:2606.17861},
  year={2026}
}

@misc{qwen2026qwen35,
  title={{Qwen3.5}: Towards Native Multimodal Agents},
  author={{Qwen Team}},
  month={February},
  year={2026},
  url={https://qwen.ai/blog?id=qwen3.5}
}

@misc{anthropic2026opus5,
  title={Introducing Claude Opus 5},
  author={{Anthropic}},
  year={2026},
  url={https://www.anthropic.com/news/claude-opus-5}
}

@misc{openai2026gpt56,
  title={{GPT-5.6}: Frontier intelligence that scales with your ambition},
  author={{OpenAI}},
  year={2026},
  url={https://openai.com/index/gpt-5-6/}
}

@misc{deepseek2026v4,
  title={{DeepSeek V4} Preview: Entering the Era of Affordable Million-Token Context},
  author={{DeepSeek}},
  year={2026},
  url={https://deepseek.com/en/news/v4-preview/}
}

@misc{zai2026glm53,
  title={{GLM-5.3} Model Card},
  author={{Z.ai}},
  year={2026},
  url={https://build.nvidia.com/z-ai/glm-5-3/modelcard}
}

@misc{qwen2026qwen38max,
  title={{Qwen3.8-Max}},
  author={{Qwen Team}},
  year={2026},
  url={https://docs.modelstudio.console.alibabacloud.com/en/model-studio/qwen3-8-max}
}

@article{kimi2026k3,
  title={Kimi K3: Open Frontier Intelligence},
  author={{Kimi Team} and others},
  journal={arXiv preprint arXiv:2607.24653},
  year={2026},
  url={https://arxiv.org/abs/2607.24653}
}

@article{abbas2023semdedup,
  title={{SemDeDup}: Data-Efficient Learning at Web-Scale through Semantic Deduplication},
  author={Abbas, Amro and Tirumala, Kushal and Simig, D{\'a}niel and Ganguli, Surya and Morcos, Ari S.},
  journal={arXiv preprint arXiv:2303.09540},
  year={2023}
}

@inproceedings{tirumala2023d4,
  title={{D4}: Improving {LLM} Pretraining via Document De-Duplication and Diversification},
  author={Tirumala, Kushal and Simig, D{\'a}niel and Aghajanyan, Armen and Morcos, Ari S.},
  booktitle={Advances in Neural Information Processing Systems},
  volume={36},
  year={2023}
}
\bibliographystyle{iclr2027_conference}

\clearpage
\appendix

\section{Qualitative Game-Generation Showcase}
\label{app:qualitative_showcase}

\begin{center}
\begin{minipage}{\textwidth}
    \centering
    \includegraphics[width=\textwidth,height=0.85\textheight,keepaspectratio]{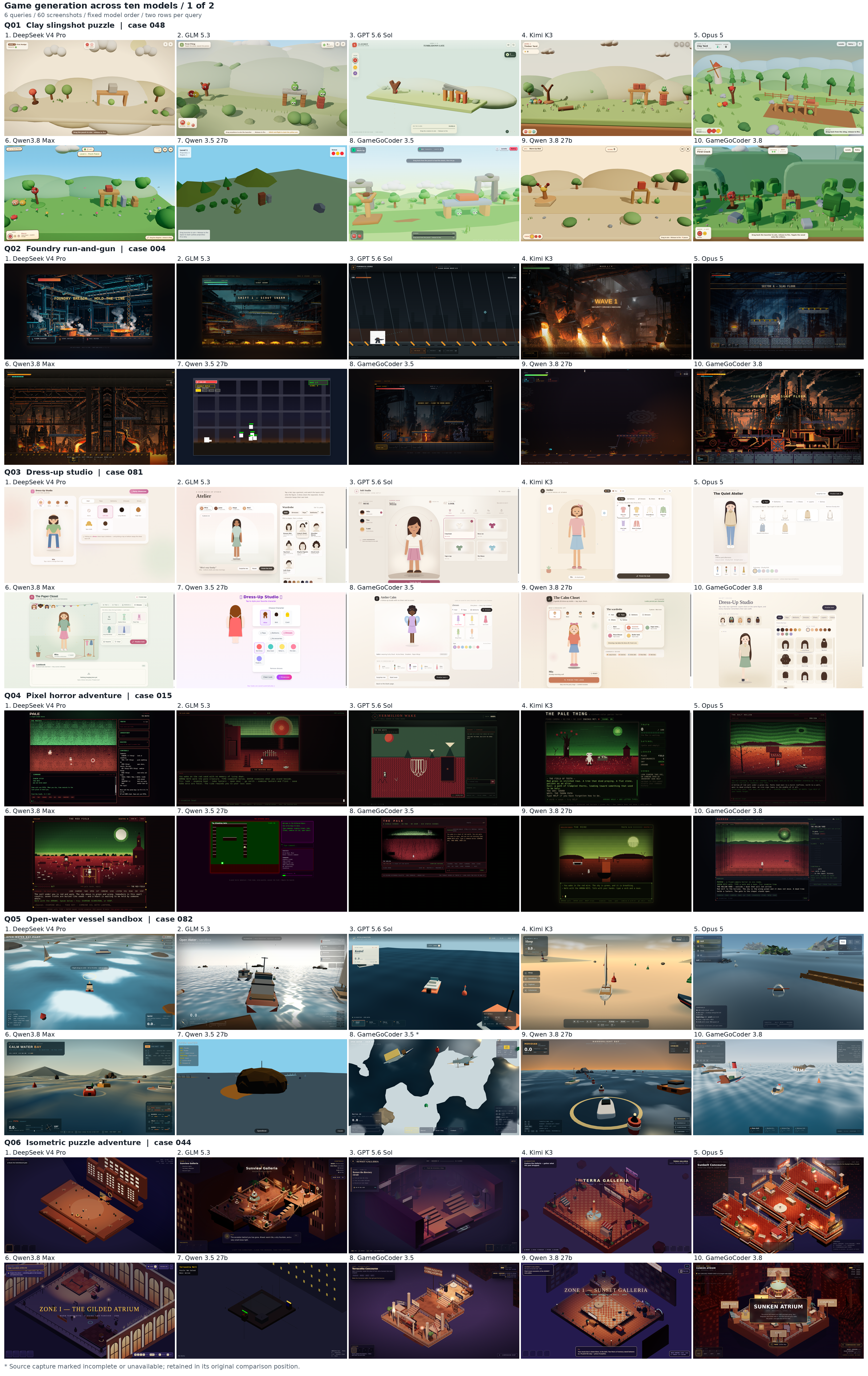}
    \begingroup
    \small
    \captionof{figure}{\textbf{Qualitative comparison of game generation across ten models (Part I).} Six tasks illustrate differences in scene composition, visual detail, and interface presentation under identical queries. Each task spans two rows, with model names labeled above the screenshots.}
    \label{fig:showcase_part1}
    \endgroup
\end{minipage}
\end{center}

\clearpage
\begin{figure}[p]
    \centering
    \includegraphics[width=\textwidth,height=0.90\textheight,keepaspectratio]{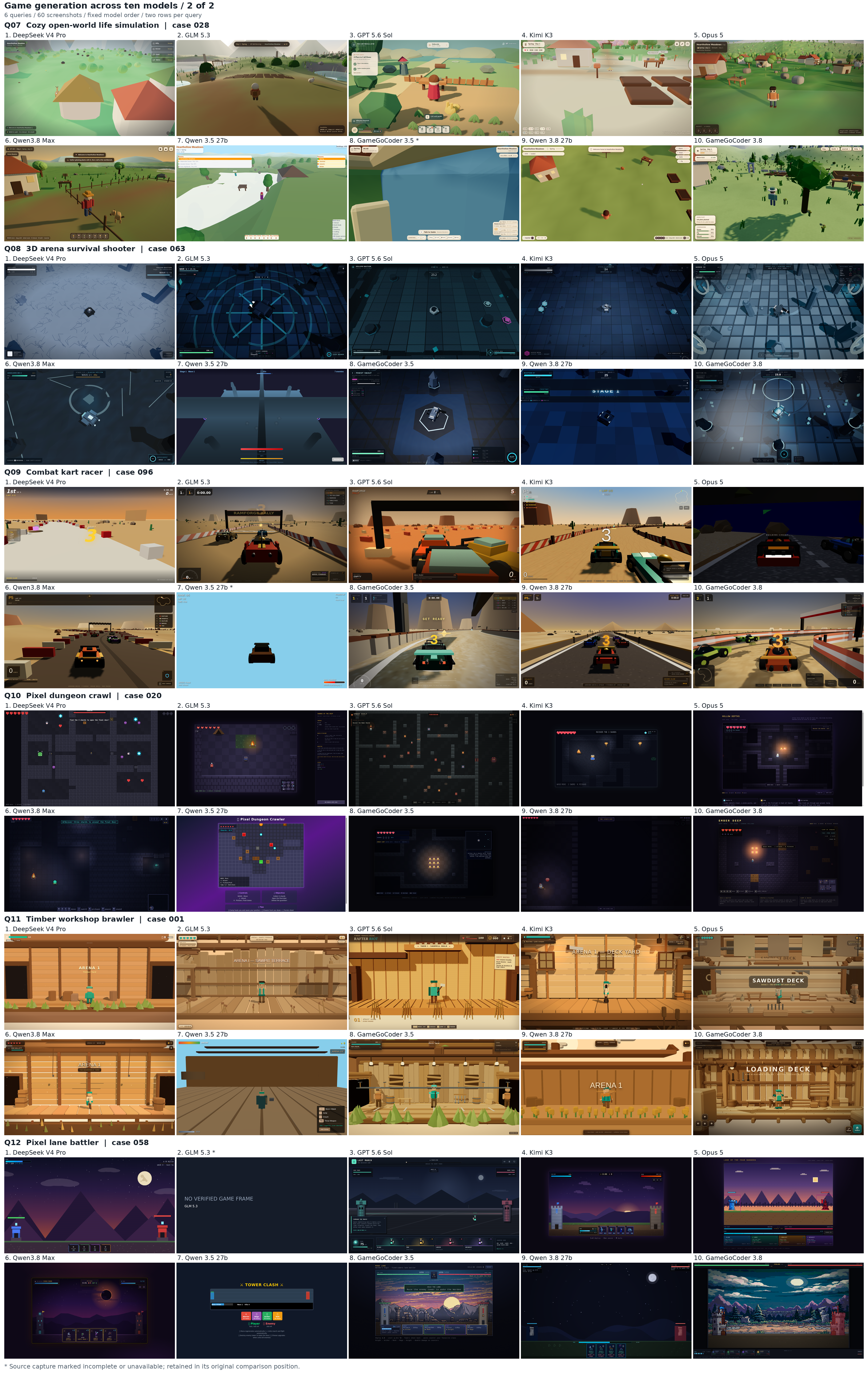}
    \begingroup
    \small
    \caption{\textbf{Qualitative comparison of game generation across ten models (Part II).} Six additional tasks extend the showcase across 2D and 3D settings, using the same model ordering as Figure~\ref{fig:showcase_part1}.}
    \label{fig:showcase_part2}
    \endgroup
\end{figure}
\clearpage

The twelve selected tasks compare ten models under identical queries. The QwenSFT labels denote GameGoCoder 3.5 and 3.8, respectively. Asterisks mark incomplete or unavailable source captures retained in their original positions. Screenshots complement the quantitative and interactive evaluations.

\section{Analytical Formulation}
\label{app:task_formulation}

Let $s$ be a game seed, $p=f(s)$ its PRD, and $q=c(p)$ the query delivered to the coding agent. The seed constrains the intended experience but generally admits multiple valid designs. The PRD describes one candidate design consistent with the seed. Its additional design choices resolve the implementation of that candidate design. We measure the subsequent compression through the remaining text length and the explicit coverage of a fixed set of gameplay relations.

Let $L(x)$ be the token length of text $x$, and let $E_s$ be a fixed inventory of gameplay clauses in $p$, such as a failure triggering respawn or a timer triggering the results screen. Each clause states a dependency together with its specified conditions and parameters. With $n_s=|E_s|>0$, let $K_s(x)$ count the clauses completely expressed in $x$. By construction, $K_s(p)=n_s$. Define retained length $r_s$, relation retention $R_s$, and omission fraction $D_s$ as
\begin{equation}
    r_s=\frac{L(q)}{L(p)},\qquad
    R_s=\frac{K_s(q)}{n_s},\qquad
    D_s=1-R_s.
    \label{eq:retention_and_omission}
\end{equation}
Here \(D_s\) measures omissions from the reference inventory. For relation density \(\rho_s(x)=K_s(x)/L(x)\), the relation density formulation yields the density gain \(G_s=R_s/r_s=(1-D_s)/r_s\). Length, retention, and density are reported together to describe the query representation.

A conceptual construction objective is to minimize query length while passing the protected-dependency checks:
\begin{equation}
    \min_q L(q)
    \quad\text{subject to}\quad
    q\in\mathcal{Q}_{\mathrm{protected}}(s,p),
    \label{eq:specification_objective}
\end{equation}
where $\mathcal{Q}_{\mathrm{protected}}(s,p)$ contains seed-compatible queries that pass the prompt-based checks for protected gameplay dependencies. Clause selection and revision implement this design objective with task-adaptive detail budgets. The separate text audit measures complete retention of sampled PRD clauses, including their exact parameters; $D_s$ describes that audit rather than an admission threshold.

We evaluate the resulting queries through the games they produce. For a fixed teacher, environment, and execution budget $B$, query $q$ induces an artifact distribution $\pi_B(a\mid q)$. If $\mathcal{V}_s$ is the set of playable implementations compatible with the seed, then
\begin{equation}
    U_B(q;s)=\Pr_{a\sim\pi_B(\cdot\mid q)}[a\in\mathcal{V}_s].
    \label{eq:valid_artifact_probability}
\end{equation}
We compute $r_s$, $R_s$, and $G_s$ for all 124 GameGoBench tasks using a fixed inventory of 927 eligible PRD clauses. Complete, partial, and absent clause labels record the level of expression in each query. Human comparisons assess artifact preference on the same tasks under the shared execution budget. Appendix~\ref{app:static_specification_analysis} details clause selection, annotation, and aggregation.

\section{Data Construction Details}
\label{app:data_construction}

\subsection{Game Seed Sources and Screening}

The data are organized into the three source families described in the main text---Steam, browser-game portals, and WeChat mini-games---and four provenance collections. The eighteen source entries comprise one Steam entry, sixteen browser-game entries across two collections, and one WeChat entry. Within the browser-game family, the legacy web-game corpus comprises historical snapshots from Y8, Playgama, 4399, Poki, and CrazyGames. The separate multi-site pool covers eleven portals, with GameDistribution, CrazyGames, ArmorGames, TwoPlayerGames, Poki, Kongregate, AddictingGames, and Kizi as its largest sources. The browser-game entry count is collection-specific: a portal represented in both collections contributes an entry to each, so these entries are not a count of unique domains. The WeChat collection comes from the public 17yoo mini-game directory, and the Steam collection comes from public store pages and metadata endpoints.

Records are mapped to a common schema while preserving field boundaries between title, description, instructions, categories, and tags. Text normalization removes HTML markup, control characters, invisible characters, and source-specific promotional templates. For the two web collections and WeChat mini games, a conservative two-stage screen checks whether a record contains concrete evidence about player actions, controlled objects, rules, objectives, feedback, or interactions. A record is removed only when both stages find no such evidence. Steam uses rules to route obvious cases and applies the same two-stage standard to boundary cases, followed by an audit of rule-based retention. Ambiguous cases and failed judgments are retained. Category and genre metadata are not used as game-type annotations.

\subsection{Gameplay-Level Deduplication}
\label{app:gameplay_dedup}

Lexical deduplication alone cannot identify games that describe the same mechanic using different themes or wording. Following prior work on semantic dataset deduplication \cite{abbas2023semdedup,tirumala2023d4}, we retrieve semantically related records, but define equivalence using a game-specific criterion: whether one core gameplay implementation and state machine can realize both records after changing only content, presentation, parameters, or optional rules.

For each record, an LLM extracts a structured gameplay signature containing the primary mechanic, core loop, player actions, controlled entities, world state, objective, scoring, failure conditions, progression, resource economy, multiplayer structure, and camera view. The extractor must ground these fields in the source text and places thematic information in a separate field. Candidate generation combines exact matching, word and character TF-IDF, MinHash, and nearest-neighbor retrieval over 1{,}024-dimensional Qwen3-Embedding-0.6B signature embeddings. We retrieve the top 15 embedding neighbors per record, merge them with lexical candidates, and use Qwen3-Reranker-0.6B to retain the top five candidates per record.

An LLM judge assigns each successfully judged candidate pair one of three relations. Relation A indicates a reusable gameplay template, relation B indicates insufficient or conflicting evidence, and relation C indicates a difference in core actions, state transitions, loops, objectives, or required failure rules. Unresolved pairs do not create duplicate edges. Only A relations create edges in the gameplay graph. Because pairwise similarity is not strictly transitive, we remove weak bridge edges that lack at least two shared A-neighbors and then apply weighted Leiden community detection. Exact matches and high-confidence edges with signature cosine similarity of at least 0.90 bypass the shared-neighbor rule.

Selection is performed within each resulting cluster. Clusters of at most five records are retained in full. For a cluster of size $n>5$, we retain $\operatorname{round}(5+\sqrt{n-5})$ records, prioritizing information sufficiency, rotation across source collections, source-text length, and finally a stable record identifier. This quota policy controls the frequency of repeated gameplay within the training pool while preserving small gameplay families and alternative source descriptions. Multiple examples of a gameplay pattern may be retained under the quota. The clustering and quota-based selection stage processes 70{,}000 records and retains 67{,}564 seeds. Training and benchmark data are strictly separated from the seed stage onward: benchmark seeds and their derived examples are excluded from all training sets.

\subsection{Specification Routes and Stage Validation}

The task-definition stage chooses the routes used by the remaining planning calls. The rendering route distinguishes layered two-dimensional interaction, true three-dimensional scenes observed from a fixed 2.5D camera, and three-dimensional worlds with a movable camera. The corresponding guidance specializes coordinate systems, input semantics, collision, camera behavior, spatial readability, and asset construction. A separate gameplay route retrieves guidance for the selected primary category and finer archetype. Small expert cards may additionally focus a planning call on applicable concerns such as combat, level flow, animation, audio, sprite production, asset sourcing, or shaders. The retrieved material is framed as attention guidance and is prohibited from adding an unsupported mechanic.

Each planning call produces a structured JSON artifact. The next call is run only after the preceding artifact can be parsed and satisfies its stage checks. The asset stage additionally validates required asset definitions, executable file types, prompt references, rendering-branch consistency, and coverage of blueprint requirements. Invalid outputs enter a bounded revision step. Canonical compilation then selects fields from the stage that owns them and removes repeated descriptions before producing the full PRD.

\subsection{Task-Adaptive Query Construction Policy}
\label{app:query_construction}

The large-scale converter first recovers the authoritative rendering dimension from affirmative declarations in the full PRD, ignoring mentions that only forbid an alternative dimension. The compressor then returns an agent-facing query together with its semantic-complexity estimate and five audit fields: preserved axes, protected risks, implementation freedom, complexity basis, and omitted details. Separate continuous output envelopes are used for two-dimensional and spatial games. Independent LLM validators check query fidelity to protected gameplay dependencies and the implementation freedom left to the coding agent. If a query falls outside its envelope, fails either semantic check, omits a protected-risk analysis, or leaves no implementation decision open, it is revised without changing the original complexity estimate.

The production version also introduces sparse implementation diversity after the task has passed a semantic eligibility test. A stable hash of the task identifier selects at most one optional direction from four cases: reflections for water-dominant 3D scenes, richer geometry for an identity-critical 3D focal object, coherent sprite sheets for eligible 2D side-view action games, or portraits for character-driven card games. The suggestion is expressed as an option rather than a task requirement. When a selected technique is adopted, a short adoption contract prevents degenerate use, such as generating a sprite sheet while rendering a procedural fallback as the default character. The selection and retention decision are stored in the audit record.

\subsection{React trajectory quality checks}
\label{app:react_quality_checks}

The curation rules target React development trajectories. As described in the main text, admission requires executable projects with complete codebases, valid parsed structures, and verified runtime success. The trajectory-level checks below form part of this multi-stage filtering protocol. Structural checks reject malformed query or tool records, undefined tool names, session timeouts, unfinished final tool-call turns, and trajectories with fewer than eight messages. Non-grep tool calls are checked for missing required arguments and undeclared parameters. Retained trajectories must contain at least one \texttt{create\_file} or \texttt{edit\_file} operation. The checks also reject repeated or malformed tool-output patterns, including concatenated \texttt{list\_files} responses and truncated JSON arguments.

Execution checks reject trajectories with at least ten failed tool calls. They also reject nominally successful tool responses that contain transport failures, failed MCP calls, or broken-pipe errors. Calls to Supabase tools must follow initialization. To limit unsuccessful repair loops, trajectories are rejected when lint failures lead to more than three file-repair rounds or when at least fourteen consecutive assistant turns edit the same file.

The build check inspects the final four assistant turns. If it finds a \texttt{build\_project} call, the corresponding result must be present and report JSON status \texttt{success}. A trajectory is not rejected solely because no build call occurs in this window. After a successful terminal build, trailing turns are removed while retaining all parallel tool results associated with that build. These checks assess the recorded development process; they complement the project-level executability, integrity, and runtime-success requirements for admission.

The retained records are serialized directly into MS-SWIFT \texttt{messages} JSONL for supervised training. Requirements and Quality scores belong to artifact evaluation and are not admission thresholds for these trajectory checks. Corpus accounting records query submission, trajectory collection and curation, and the model-specific training counts in Table~\ref{tab:corpus_accounting}.

\subsection{Corpus accounting}
\label{app:corpus_accounting}

To this end, from 67{,}564 submitted task-specific queries, 67{,}487 initial trajectories are recorded. Among these, 65{,}082 complete the primary execution phase successfully, achieving a 96.3\% completion rate over submitted queries. Subsequent artifact integrity and formatting verification retain 55{,}060 examples, corresponding to an overall GameGoData yield of 81.5\% over submitted queries. Table~\ref{tab:corpus_accounting} summarizes the seed-selection, collection, curation, and training counts. The 55{,}060 curated examples constitute the final GameGoData corpus and are used as input to both dense-model training runs; model-specific length filtering removes no further examples.

\begin{table}[t]
\centering
\caption{Seed-selection, trajectory collection, and model-specific training counts.}
\label{tab:corpus_accounting}
\small
\begin{tabular}{@{}lr@{}}
\toprule
Seed-selection, collection, or training stage & Records \\
\midrule
Records processed by clustering and quota-based selection & 70,000 \\
Seeds retained after quota-based selection & 67,564 \\
Task-specific queries submitted & 67,564 \\
Initial trajectories recorded & 67,487 \\
Primary execution completed successfully & 65,082 \\
Final GameGoData after integrity and formatting checks & 55,060 \\
\midrule
Dense-model training input before length filtering & 55,060 \\
Qwen3.5-27B after length filtering & 55,060 \\
Qwen3.8-27B after length filtering & 55,060 \\
\bottomrule
\end{tabular}
\end{table}

\subsection{GameGoBench Task Selection}
\label{app:benchmark_split}

GameGoBench seeds were selected and separated from the training seeds before development of the task-construction pipeline, using stratified selection to cover rendering formats and implementation difficulties. Training and test data are strictly isolated from the seed stage onward: none of the 124 benchmark seeds, their derived specifications, queries, or trajectories enters any training set. The benchmark contains 47 2D, 22 2.5D, and 55 3D tasks, stratified into Easy, Medium, and Hard within each rendering format. The training and test seeds subsequently pass through the same specification and query-construction pipeline, which was not designed around the test set. Details of gameplay clustering and training-pool quotas are provided in Appendix~\ref{app:gameplay_dedup}.

\section{Evaluation Design and Collection Cost}
\label{app:evaluation_design}

The study evaluates task representations through fixed-teacher artifact comparisons and corpus utility through student fine-tuning. Direct queries are produced by DeepSeek-V4-Flash-Vision-Exp from source seeds, rewriting supported gameplay into self-contained browser-game requests without the multi-stage PRD. All 124 query identifiers match the PRD and task-specific conditions. Their median length is 227.5 tokens, close to 235 for task-specific queries; these are seed rewrites rather than PRD summaries. The held-out comparison evaluates the construction procedure without tuning it on GameGoBench.

Collecting high-quality teacher trajectories is the main cost of representation-level training comparisons. Each alternative query changes implementation decisions, tool interactions, execution feedback, and repairs, requiring a fresh sandbox rollout and subsequent student training. Repeating this process for direct queries and full PRDs at corpus scale would multiply the teacher-collection cost. We therefore compare all three representations through controlled artifact evaluations on the 124 held-out seeds, while concentrating large-scale trajectory collection and SFT on the task-specific queries. This design allocates teacher compute to the final training corpus and evaluates the representation choices under a shared teacher, scaffold, and interaction budget. The same confidential teacher is used for trajectory collection and the task-representation comparisons. In each comparison condition, the task input is the corresponding direct query, full PRD, or task-specific query; the compact condition does not additionally supply the full PRD or the three planning-stage outputs. The Base/MoE study additionally examines learning from a pretrained initialization.

The static analysis covers the same 124 GameGoBench tasks as the artifact comparisons. Paired text lengths measure compression, while an audit of 927 PRD clauses measures complete, partial, and absent expression in the task-specific queries. We report clause retention and density alongside artifact preferences on this shared task set. Downstream experiments evaluate the training utility of the collected corpus across model backbones.

\section{Training and Public Evaluation Details}
\label{app:experimental_details}

\subsection{Model and Inference Settings}
\label{app:model_settings}

The principal dense models are Qwen3.5-27B and Qwen3.8-27B, initialized from their corresponding instruction-tuned checkpoints. A separate Qwen3.5-35B-A3B MoE run is retained as an additional training study. All three runs apply SFT to all language-model parameters rather than using LoRA, while visual towers and aligners remain frozen. The final GameGoData corpus contains 55{,}060 trajectories, all of which are retained for both dense-model training runs after length filtering. The dense-model configurations and training counts are reported below.

The local training runs use NVIDIA H800 GPUs with approximately 80 GB per GPU, BF16 arithmetic, public random seed 42, and Linux. The dense runs use PyTorch 2.9.1+cu128, MS-SWIFT 4.4.0, Transformers 5.12.1, DeepSpeed 0.19.2, Megatron-Core 0.16.1, FlashAttention 2.8.3, and Transformer Engine 2.10.0. The training trajectories are converted directly into MS-SWIFT \texttt{messages} JSONL records with \texttt{template=qwen3\_5}, \texttt{agent\_template=qwen3\_5}, and \texttt{padding\_free=true}. With \texttt{truncation\_strategy=delete}, an over-length trajectory is removed as a whole rather than left- or right-truncated.

Within each benchmark, all compared models use the same Code Arena scaffold, sandbox environment, and inference budget, as specified in the main text. For GameGoCoder and its matched base checkpoints, the inference settings are temperature $0.6$, top-$p=0.95$, at most 100 interaction steps, and a maximum of 65{,}536 generated tokens per model call. The dense models are served with vLLM in BF16 using tensor parallelism 8 and a 262{,}144-token context limit. Each model makes one generation attempt per query. The accompanying configurations specify training and inference settings.

\subsection{Dense-model training configuration}
\label{app:dense_training}

Each dense model is trained for one epoch over its retained GameGo training subset. After length filtering, both the Qwen3.5-27B and Qwen3.8-27B runs contain 55{,}060 training trajectories. Both use fused AdamW, peak learning rate $1\times10^{-5}$, cosine decay to zero, 3\% warm-up, $(\beta_1,\beta_2)=(0.9,0.95)$, $\epsilon=10^{-8}$, weight decay $0.1$, gradient clipping at $1.0$, gradient checkpointing, and DeepSpeed ZeRO-3 with CPU-pinned optimizer offload. The effective global batch size is 8 and the maximum supported sequence length is 262{,}144 tokens.

\begin{center}
\scriptsize
\setlength{\tabcolsep}{0pt}
\begin{tabular}{@{}p{0.20\linewidth}p{0.16\linewidth}p{0.18\linewidth}p{0.18\linewidth}p{0.20\linewidth}@{}}
\toprule
Model & GPUs & Training input $\rightarrow$ retained & Sequence / packing & Final checkpoint \\
\midrule
Qwen3.5-27B & 8 or 16 & 55,060 $\rightarrow$ 55,060 & up to 262,144; packing & final dense checkpoint \\
Qwen3.8-27B & 32 & 55,060 $\rightarrow$ 55,060 & up to 262,144; packing & final dense checkpoint \\
\bottomrule
\end{tabular}
\end{center}

The Qwen3.5 dense runs use 8 or 16 H800 GPUs across launches, reflecting resource allocation. Activation CPU offload supports long-context training. Checkpoints contain model, optimizer, scheduler, and RNG state. The final reported model is the last checkpoint of its corresponding SFT run.

\subsection{MoE training study}
\label{app:moe_training}

The Qwen3.5-35B-A3B Base experiment uses 64 H800 GPUs with MS-SWIFT Megatron/Megatron-Core rather than DeepSpeed. Training runs for six epochs at sequence length 262,144 with 19,148 packed sequences. Tensor, pipeline, data, and expert parallelism are TP=8, PP=1, DP=8, and EP=8, with expert tensor parallelism ETP=1, context parallelism CP=1, and sequence parallelism enabled. The microbatch size is 1 and the global batch size is 128, corresponding to 16 microbatches per optimizer update on each data-parallel replica. The model has 256 routed experts and activates the top 8 experts per token.

The MoE run uses Adam with a linear learning-rate decay from $5\times10^{-5}$ to $5\times10^{-6}$, no warm-up, $(\beta_1,\beta_2)=(0.9,0.95)$, $\epsilon=10^{-8}$, weight decay $0.1$, and gradient clipping at $1.0$. Router computation uses FP32 with load-balancing coefficient $10^{-3}$; grouped GEMM, all-to-all dispatch, FlashAttention, fused cross entropy, and full uniform activation recomputation are enabled. Epoch checkpoints are checkpoint-149, -298, -447, -596, -745, and -894, with checkpoint-897 as the training endpoint. This run resumed from weights without restoring optimizer moments or RNG state. Restarted jobs preserve model weights and scheduler position and initialize fresh optimizer state.

\subsection{Loss masking and serialization}
\label{app:loss_masking}

The response-masked objective in the main text is implemented at message level. \texttt{system}, \texttt{user}, and \texttt{tool\_response} tokens receive label $-100$ and contribute context only. Assistant thinking, assistant natural-language responses, and tool calls are supervised with their original token IDs. Tool calls are first serialized as assistant-side messages, whereas tool responses are serialized as \texttt{tool} messages. Consequently, the objective supervises the agent's decisions and actions while preserving execution feedback as conditioning context.

\subsection{Public Benchmark Evaluation}
\label{app:public_benchmark}

We evaluate 413 ArtifactsBench-G tasks and 156 Cookie-Bench-G tasks using their official queries and evaluation pipelines, with Claude Sonnet 4.6 powering both verifier pipelines as in the main text. Each model generates one artifact for every query. Failed builds receive the score prescribed by the corresponding official evaluator. Table~\ref{tab:main_results} reports task-level mean scores; comparisons with the corresponding base checkpoint use the same benchmark tasks. The reported values are task-level mean point estimates.

\subsection{GameGoBench Verifier Sensitivity}
\label{app:verifier_scope}

The main results use Claude Sonnet 4.6, DeepSeek-V4-Flash-Vision-Exp and Qwen3.8-Max provide additional evaluations in Table~\ref{tab:verifier_sensitivity}. Each verifier configuration runs the two-stage protocol described in the main text: an interactive multimodal agent plays the game in a headless browser under step-driven execution, and a separate evaluation agent scores the resulting interaction trace. Each configuration collects its own observations. Requirements scores cover core concepts, input responsiveness, game loops, and query specifications; Quality scores use chronologically sampled frames to assess subject art, scene composition, and UI integration. Every model attempts 124 tasks, yielding $N_{\mathrm{repo}}$ React repositories, with counts from 117 to 124. Of these, $N_{\mathrm{render}}$ display a rendered page. The Execution Rate reported as Exec. in Table~\ref{tab:main_results}, called Render in the evaluation records, is $100N_{\mathrm{render}}/N_{\mathrm{repo}}$. Tasks without a repository are excluded from this denominator; generated repositories that fail to render remain in it. This repository-conditioned rate measures rendering success; runtime behavior is additionally examined in the interactive traces. Repository availability and rendering success are separate stages. The three verifier configurations use the same generated repository sets and have identical Exec. scores. The main table retains Sonnet 4.6 results.

For verifier $j$, the aggregate Overall score is $S_j=(\mathrm{Req}_j+Q_j)/2$, where $\mathrm{Req}_j$ and $Q_j$ are the aggregate Requirements and Quality scores. For each evaluated model, we report the cross-verifier mean $\bar{S}=\frac{1}{3}\sum_{j=1}^{3}S_j$ and sample variance $s^2=\frac{1}{2}\sum_{j=1}^{3}(S_j-\bar{S})^2$. These statistics summarize variation among the three verifier-level aggregate scores.

\begin{table}[t]
\centering
\caption{GameGoBench Overall scores under three verifiers. DS-Flash denotes DeepSeek-V4-Flash-Vision-Exp and Qwen denotes Qwen3.8-Max. Mean and variance are computed across the three verifiers from unrounded component scores; variance is in squared score points.}
\label{tab:verifier_sensitivity}
\small
\setlength{\tabcolsep}{4pt}
\begin{tabular}{@{}lrrrrr@{}}
\toprule
Evaluated model & Sonnet 4.6 & DS-Flash & Qwen & Mean & Variance \\
\midrule
Claude Opus 5 & 61.84 & 61.88 & 61.66 & 61.79 & 0.0139 \\
GPT-5.6 & 52.78 & 51.46 & 54.04 & 52.76 & 1.6584 \\
DeepSeek-V4-Pro & 49.52 & 50.70 & 49.44 & 49.89 & 0.4992 \\
GLM-5.3 & 59.81 & 60.18 & 60.79 & 60.26 & 0.2447 \\
Qwen3.8-Max & 58.35 & 58.29 & 57.99 & 58.21 & 0.0374 \\
Kimi-K3 & 53.82 & 53.73 & 53.48 & 53.67 & 0.0294 \\
Qwen3.5-27B & 34.00 & 33.73 & 33.78 & 33.84 & 0.0209 \\
GameGoCoder 3.5 & 51.85 & 51.80 & 52.69 & 52.11 & 0.2514 \\
Qwen3.8-27B & 52.53 & 52.39 & 52.02 & 52.31 & 0.0690 \\
GameGoCoder 3.8 & 55.29 & 54.16 & 54.68 & 54.71 & 0.3213 \\
\bottomrule
\end{tabular}
\end{table}

The DeepSeek-V4-Flash-Vision-Exp and Qwen3.8-Max rankings have Spearman correlations of 0.964 and 0.976 with the Sonnet 4.6 ranking, respectively; their mean absolute Overall-score differences from Sonnet are 0.47 and 0.54 points. GameGoCoder 3.5 improves over its matched baseline by 17.85, 18.06, and 18.91 points under Sonnet, DeepSeek, and Qwen, respectively; the corresponding gains for GameGoCoder 3.8 are 2.76, 1.78, and 2.66 points. Thus, both matched-baseline improvements persist across all three verifiers, and the aggregate rankings remain similar. The three model families provide complementary automated evaluations with closely aligned aggregate rankings. The blinded human comparisons provide complementary evidence.

\subsection{GameGoBench Trajectory Analysis}
\label{app:gamegobench_trajectory}

The trajectory analysis described in Figure~\ref{fig:main_trajectory_analysis} uses the same 124 GameGoBench query identifiers for each included model. Trajectories are aligned by \texttt{data\_id}, not by file order, and the three query groups contain 13, 40, and 71 tasks. We aggregate all assistant-generated content over turns for trajectory length, and report thinking length separately. Interaction steps count assistant turns with their resulting tool observations; simultaneous tool calls are retained as one step and their width is recorded independently. Build statistics count calls to \texttt{build\_project} and record the first and last build positions both in absolute steps and relative trajectory quartiles.

For each model and metric, the supplementary analysis reports p10, p25, p50, p75, p90, p95, and p99 quantiles together with the interquartile range. Tool reliability is summarized at both the individual-call level and the per-trajectory level, while infrastructure failures are retained as a separate category rather than being silently treated as model errors. Most providers expose completion and reasoning usage directly; when usage is unavailable, we estimate the corresponding lengths with a common Qwen tokenizer. Those estimates support relative behavioral comparisons under a common tokenization scheme. The full per-query, tool-distribution, build-position, and group-stratified statistics are released as supplementary analysis artifacts.

\section{Human Pairwise Evaluation}
\label{app:human_evaluation}

The same six hired independent expert evaluators assess available playable artifact pairs from the 124 benchmark tasks; pair availability depends on the compared systems. For the model comparisons in Figure~\ref{fig:pairwise_preference}, both GameGoCoder variants are compared with the corresponding comparison models with model identities hidden. Each reviewer plays every eligible game from its start through the end of the playable experience before making a pairwise judgment. Sessions follow completion of the game rather than a fixed number of minutes. Reviewers select the first artifact, the second artifact, or a tie. The task-representation study reports 124 artifact pairs for each of its three comparisons. Reviewers assess fidelity, playability, and overall quality against the original seed. Table~\ref{tab:pipeline_ablation} pools their judgments across artifact pairs. Its win, tie, and loss counts are reviewer judgments rather than counts of distinct tasks, and the percentages use $W+T$ as the denominator. Full PRDs receive 69.0\% preference against direct queries, while task-specific queries receive 85.7\% against full PRDs and 86.6\% against direct queries.

For the MoE checkpoint comparisons, all six reviewers assess each eligible artifact pair with checkpoint identities hidden. Table~\ref{tab:moe_pairwise_ablation} includes only pairs for which both games are playable. Its $N$ values of 110, 108, and 123 count artifact pairs, while W/T/L pools reviewer judgments over those pairs. Epoch~2 receives preference scores of 0.852 against Base and 0.664 against the official post-SFT MoE checkpoint. Epoch~6 receives 0.807 against the official checkpoint. Each comparison uses its corresponding set of eligible playable pairs.

Both tables report W/T/L from the first system's perspective and compute preference as $W/(W+L)$, excluding ties. Each eligible artifact pair receives six judgments, which are pooled to obtain the reported preference. The MoE comparisons use the playable pairs listed in Table~\ref{tab:moe_pairwise_ablation}. Reviewer agreement is reported separately below.

\paragraph{Inter-reviewer agreement.}
We report raw pairwise agreement among the six reviewers. For each evaluated artifact pair $i$, let $n_{iW}$, $n_{iT}$, and $n_{iL}$ denote the numbers selecting a win, tie, and loss for the first system, with $n_{iW}+n_{iT}+n_{iL}=6$. The fraction of the 15 reviewer pairs giving the same judgment is
\begin{equation}
    A_i=\frac{\sum_{c\in\{W,T,L\}}n_{ic}(n_{ic}-1)}{6\cdot5}.
    \label{eq:reviewer_raw_agreement}
\end{equation}
Ties are retained as a separate judgment category. Averaged over the evaluated artifact pairs within each study, raw agreement is 82.5\% for the model comparisons in Figure~\ref{fig:pairwise_preference}, 81.7\% for the task-representation comparisons in Table~\ref{tab:pipeline_ablation}, and 84.7\% for the MoE checkpoint comparisons in Table~\ref{tab:moe_pairwise_ablation}. Each percentage summarizes the corresponding evaluation as a whole rather than every individual comparison row. These percentages report observed reviewer agreement across the three studies.

\section{GameGoBench Task-Representation Analysis}
\label{app:static_specification_analysis}

\subsection{Task matching and length measurement}

We analyze the stored PRDs and task-specific queries of all 124 GameGoBench tasks used in the artifact comparisons. Records are aligned by their original task identifiers across the PRD file and the query exports containing 40, 13, and 71 tasks. Three empty query fields in the exports are recovered from their archived gallery inputs and checked against the corresponding original query records. The analysis retains the fixed evaluation queries and records source hashes and recovery provenance.

Token lengths are measured with the Qwen3.5-27B tokenizer, with special tokens disabled and no chat template. We count the stored Markdown and serialized specifications. The first complete \texttt{Deterministic Asset Execution Contract} heading separates each PRD into its body and asset appendix. We compute retained length and compression for each matched task before aggregating. Table~\ref{tab:bench_text_audit} reports medians of these per-task ratios alongside median text lengths.

\subsection{Clause selection and annotation}

Reference clauses are selected from PRD text before examining query coverage. A deterministic script extracts spans of 45 to 230 characters containing a condition or transition cue, prioritizing core mechanics and rule sections and using controls, progression, and other gameplay sections as a fallback. Asset, execution, feedback, color, shader, performance, and acceptance sections are excluded. Candidates are ordered by the SHA-256 hash of the task identifier and clause text, and up to eight are selected per task. The inventory is frozen before semantic annotation, with exact source paragraphs and clause hashes retained.

A single AI-assisted text audit reviews the 977 candidates and excludes 50 attribute lists, presentation or implementation requirements, and uninterpretable fragments, leaving 927 gameplay clauses across all 124 tasks. Each eligible clause is labeled \emph{complete} when the query expresses the full dependency, including its stated conditions and parameters; \emph{partial} when the mechanism is expressed with some conditions, values, or consequences unspecified; or \emph{absent} when the dependency is unexpressed or contradicted. Complete and partial labels include exact query quotations. The scoring script checks source and query quotations, task alignment, and arithmetic. These semantic annotations are AI-assisted; the six independent expert evaluators separately judge the generated games.

For task $s$, the reference inventory $E_s$ contains its $n_s$ eligible clauses, and $K_s(q)$ counts the clauses labeled complete. We compute $R_s=K_s(q)/n_s$, $D_s=1-R_s$, and $G_s=R_s/r_s$ using the definitions in Appendix~\ref{app:task_formulation}. Partial clauses are reported as a separate category, so $D_s$ includes both partial and absent clauses. Each reference clause is weighted equally. We report pooled clause percentages and medians of per-task coverage and density separately; all 124 tasks contribute to the medians, including tasks with zero complete coverage.

\subsection{Compression, coverage, and density}

Median PRD and query lengths are 9{,}136 and 235 tokens. The median retained-length fraction is 2.26\%, and median paired compression is $44.2\times$; compression relative to the PRD body is $21.6\times$. Every query contains less than 5\% of its full PRD's tokens. Across the 927 eligible clauses, 112 are complete, 577 are partial, and 238 are absent. Median complete-clause coverage is 12.5\%, and the median density gain is $3.79\times$ against full PRDs and $1.78\times$ against PRD bodies.

\begin{table}[t]
\centering
\caption{Text analysis of all 124 GameGoBench tasks. Lengths, coverage, and density gains are per-task medians; clause percentages pool the 927 eligible clauses. Compression is the median of paired token-count ratios. Complete coverage includes the clause's stated conditions and parameters.}
\label{tab:bench_text_audit}
\small
\begin{tabular}{@{}lr@{}}
\toprule
Measurement & Result \\
\midrule
Full PRD / body / query length in tokens & 9,136 / 4,314.5 / 235 \\
Retained length, full PRD & 2.26\% \\
Compression, full PRD / body & $44.2\times$ / $21.6\times$ \\
\midrule
Complete clauses & 112 / 927 = 12.1\% \\
Partial clauses & 577 / 927 = 62.2\% \\
Absent clauses & 238 / 927 = 25.7\% \\
Per-task complete coverage $R_s$ & 12.5\% \\
Density gain $G_s$, full PRD / body & $3.79\times$ / $1.78\times$ \\
Tasks with $G_s>1$, full PRD & 68 / 124 \\
Tasks with zero complete clauses & 56 / 124 \\
\bottomrule
\end{tabular}
\end{table}

The clause labels distinguish preserved mechanisms from complete specification detail. For example, the workshop-brawler query names an airborne dive strike, while the sampled PRD clause additionally specifies its velocity, damage, impact radius, and target-height restriction; this clause is labeled partial. The 56 tasks with no complete sampled clause contribute zero density gain. These text measurements characterize the same tasks whose generated games are compared in Table~\ref{tab:pipeline_ablation}, connecting the analysis of execution requirements with artifact-level evaluation.

\section{Task-Construction Compilers and Prompt Specifications}
\label{app:harness_prompts}

This appendix describes the implementation of the two pipelines used to construct tasks. The complete prompt specifications and routing policy will be released with the accompanying implementation upon acceptance, subject to applicable software licenses, as stated in the reproducibility statement. This release does not include GameGoData trajectories or their associated artifacts. The listings below record the canonical stage interfaces and the decisions that affect downstream execution. Both pipelines use three planning stages followed by canonical PRD compilation and task-adaptive query construction; downstream implementation executes the resulting query. They differ in the source envelope and the visual-asset contract.

\subsection{End-to-end pipeline}
\label{app:harness_pipeline}

The Image-Grounded Game Task Compiler accepts either an ordinary query or a record containing minimal text fields and at most five screenshots. The Text-Grounded Web-Game Task Compiler accepts an ordinary query or a web-game catalog record containing title, description, instructions, categories, and tags; it does not read screenshots or source URLs. In both cases, the execution order is:

\begin{center}
\small
\begin{tabular}{p{0.18\linewidth}p{0.29\linewidth}p{0.42\linewidth}}
\toprule
Stage & Stage output & Function and downstream use \\
\midrule
01 & Seed specification & Distill source facts, remove reference identity, classify gameplay and rendering routes, determine scope, and define locked requirements. \\
02 & Production blueprint & Resolve the authoritative gameplay flow, rules, controls, scenes, camera, UI, feedback, technology, and acceptance criteria. \\
03 & Asset contract & Ground every required scene and interaction in executable visual, audio, procedural, image, or shader assets and stable integration locations. \\
Canonical + Compact & Full PRD and task-specific query & Compile authoritative stage fields into a full PRD, then preserve protected dependencies in a concise English query. \\
Execution & Game artifact and trajectory & Execute the task-specific query in the downstream sandbox and record the development trajectory and runnable artifact. \\
\bottomrule
\end{tabular}
\end{center}

In unattended batch production, both pipelines stop after Canonical + Compact and write one task-identifier/query record per input; execution is then run by the downstream trajectory-collection process. The three planning-stage outputs and the full PRD are internal construction representations; the task-specific query is the task description supplied for GameGoData collection. Each stage must emit its declared output before the next stage starts. The default revision budget is one revision per stage. Invalid JSON, failed schema checks, or failed asset-contract checks trigger this bounded revision; unresolved failures are retained as failed items rather than silently exported.

\subsection{Stage 1: seed distillation and routing}
\label{app:stage1_routing}

Stage 1 is the only stage that interprets the source record. It emits a self-contained, de-identified specification and does not prescribe the implementation architecture. The routing decisions are authoritative for all later stages:

\begin{itemize}[leftmargin=1.4em]
\item \textbf{Gameplay route:} exactly one primary category from the 20-category taxonomy, one finer-grained \texttt{gameplay\_archetype}, and only implementation-relevant secondary tags. The output also lists archetype-specific required phases and a complete \texttt{gameplay\_flow\_contract} from entry to restart.
\item \textbf{Rendering route:} \texttt{2d} for layered Canvas/SVG/Sprite interaction, \texttt{2.5d} for real 3D geometry under a fixed camera, and \texttt{3d} when play requires camera motion or free spatial observation. \texttt{camera\_mobility} and concrete \texttt{dimension\_evidence} must agree with this choice.
\item \textbf{Scope route:} \texttt{compact\_loop} for a short fixed round and \texttt{multi\_phase\_journey} for non-interchangeable setup, progression, exploration, or multi-scene phases. The route is based on the required player journey, not on asset count or visual style.
\item \textbf{Asset route:} \texttt{procedural\_pixel} is selected only when the source contains explicit pixel-art evidence; otherwise the default is \texttt{generated\_hybrid}. The evidence is copied into \texttt{asset\_production\_evidence}.
\end{itemize}

For Steam inputs, screenshots are used only to record reliable \texttt{image\_only\_facts}; uncertain visual inferences are separated and the source title, characters, locations, trademarks, and text are removed from the downstream request. For WebGame inputs, the visual plan is text-derived. The text-grounded pipeline therefore additionally emits \texttt{visual\_evidence\_level}, semantic visual anchors, and separate environment/entity/UI/typography color roles. These fields constrain later visual choices rather than adding unsupported style identity.

\begin{tcolorbox}[enhanced,breakable,colback=white,colframe=black,title=\textbf{Prompt: Stage-1 Task Definition},fonttitle=\small]
\begin{Verbatim}[fontsize=\scriptsize]
You are Stage 1: Task Definition.
Input: an ordinary game request, or a source record with minimal text
fields, with up to five screenshots for Steam.

Extract transferable gameplay and visual facts; remove reference identity;
do not write code or reproduce the source game. Return exactly one structured
seed specification containing:
  source facts and locked requirements;
  one primary gameplay category, one gameplay archetype, and secondary tags;
  genre-required phases and the complete flow from entry to restart;
  game_dimension/rendering_branch, camera_mobility, and dimension evidence;
  compact_loop or multi_phase_journey with a reason;
  asset_production_mode and its source evidence;
  core mechanics, actions, goal, progression, scene, HUD, feedback, and
  a self-contained de-identified prd_request.

The rendering branch and the gameplay route are authoritative for Stages 2 and 3.
Do not add mechanics unsupported by the source.
\end{Verbatim}
\end{tcolorbox}

\subsection{Stages 2 and 3: conditional planning and asset grounding}
\label{app:stage23_routing}

Stage 2 consumes only the approved Stage-1 output and the de-identified source context. The pipeline loads a rendering overlay corresponding to the selected rendering branch and a gameplay profile selected by the primary category and archetype. The gameplay router always provides a core gameplay lens; it may additionally provide combat, level-flow, game-feel, or audio guidance when the approved seed contains the corresponding evidence. These lenses focus attention on failure-prone implementation relations and are explicitly forbidden from introducing new product requirements.

The deterministic skill policy is shared by the two pipelines. Stage 1 receives the routing-observer lens. Stage 2 starts with \texttt{gameplay\_core}; it adds \texttt{action\_combat} for combat-oriented evidence, \texttt{level\_flow} for spatial categories, 3D tasks, or multi-phase journeys, \texttt{animation\_game\_feel} for action/arcade/rhythm-like categories, and \texttt{audio\_web} when audio or mood is supported by the seed. Stage 2 is capped at four lenses. Stage 3 starts with \texttt{art\_bible\_compact} and \texttt{asset\_source\_gate}; it may add \texttt{sprite\_pipeline} for eligible 2D generated-hybrid tasks and \texttt{audio\_web} when needed. The text-grounded pipeline additionally adds \texttt{shader\_fx} only when Stage 2 emits a concrete shader opportunity, and injects at most three technique metadata references. Skill selection is recorded per run and can be disabled for ablations.

Stage 2 owns the canonical gameplay representation: \texttt{gameplay\_flow\_contract}, rules, controls, entities, progression, scene/state flow, camera and viewport contracts, feedback, implementation technology, and observable acceptance criteria. Stage 3 reads the approved seed and blueprint, loads the same rendering and gameplay routes, and owns only the executable asset plan. It maps each required scene, entity, UI element, transition, and feedback to a stable asset ID, source, dimensions, generation prompt or procedural construction, integration location, and scene coverage record. The text-grounded asset contract additionally supports project-owned inline shaders. Shader guidance is injected only when the blueprint contains an approved shader opportunity; at most three compact technique references are supplied, and the generated module must define its own uniforms, render state, performance bound, gameplay variation, and fallback.

\begin{tcolorbox}[enhanced,breakable,colback=white,colframe=black,title=\textbf{Prompt: Stage-2 Game Production Blueprint},fonttitle=\small]
\begin{Verbatim}[fontsize=\scriptsize]
Read the approved seed specification. Preserve its rendering branch, gameplay
category/archetype, required phases, scope profile, visual evidence boundary,
and asset-production mode.

Produce exactly one production blueprint. Resolve the executable game:
camera and viewport; rules and ordered state flow; controls, entities,
progression, win/lose/restart; scenes, screens, UI, feedback, technology,
performance, and acceptance criteria. Keep gameplay_flow_contract as the sole
authoritative end-to-end order, and make every scene/state transition reachable.
Use rendering-specific and gameplay-specific guidance only where supported by
the seed. Do not add an unrequested mechanic or change the rendering branch.
\end{Verbatim}
\end{tcolorbox}

\begin{tcolorbox}[enhanced,breakable,colback=white,colframe=black,title=\textbf{Prompt: Stage-3 Asset Contract},fonttitle=\small]
\begin{Verbatim}[fontsize=\scriptsize]
Read the seed specification and production blueprint. Produce exactly one
asset contract that is executable by the implementation stage.

Cover every required scene, entity, UI element, transition, and feedback with
stable asset IDs, source type, integration location, dimensions, generation or
procedural construction, integration requirements, and scene coverage.
Preserve the approved rendering branch, visual anchors/color roles, and asset
production mode. For 2.5D/3D, specify geometry, materials, lighting, motion,
and bounded scene density. For inline shaders, specify uniforms, render state,
performance budget, gameplay-triggered variation, and a fallback; never copy
reference shader source. Leave uncovered_blueprint_requirements empty.
\end{Verbatim}
\end{tcolorbox}

\subsection{Canonical Compilation and Task-Adaptive Query Construction}
\label{app:canonical_compact}

Canonical compilation prevents later stages from redefining earlier decisions. Stage 1 owns product identity, source-derived constraints, routing, scope, and originality rules; Stage 2 owns gameplay and implementation; Stage 3 owns visual and audio realization. The compiler selects only these authoritative fields, replaces repeated prose with stable IDs and references, and removes the Stage-1 draft request once the blueprint has resolved it. For image-grounded assets, reference identity is redacted before downstream stages and graphic planning context is neutralized; for the text-grounded pipeline, shader contracts are reduced to executable project-owned modules rather than copied reference compositions.

Task-adaptive query construction recovers the authoritative rendering dimension, estimates semantic complexity from dependencies among player actions, states, and camera behavior, and checks the effect of removing each candidate clause. It retains a clause when removing it would change task identity, break the playable loop, make controls or space ambiguous, or remove a critical visual dependency. It deliberately leaves routine architecture, exhaustive counts, ordinary layout, and nonessential tuning open to the coding agent. The output is an English query plus an audit of preserved axes, protected risks, implementation freedom, complexity basis, and omitted detail.

\begin{tcolorbox}[enhanced,breakable,colback=white,colframe=black,title=\textbf{Prompt: Task-Adaptive Query Construction},fonttitle=\small]
\begin{Verbatim}[fontsize=\scriptsize]
Given the canonical seed, blueprint, and asset contract, write one concise
English agent-facing query. Preserve identity, rendering/spatial behavior,
camera and controls, ordered gameplay dependencies, failure/recovery, and
critical visual requirements. For each candidate clause ask whether deleting
it would materially change the implemented game; retain only protected
dependencies and leave routine implementation choices open.

Return the compact query and an audit containing preserved axes, protected
risks, implementation freedom, complexity basis, and omitted details. Do not
invent mechanics, repeat stage-owned prose, or change the rendering branch.
\end{Verbatim}
\end{tcolorbox}

\subsection{Validation and implementation handoff}
\label{app:harness_validation}

The pipeline materializes structured outputs only after successful parsing and object-level checks. Required top-level fields are checked per stage; rendering branch consistency and cross-stage field ownership are enforced; Stage 3 additionally checks executable extensions, asset-source contracts, required asset IDs, prompt references, and complete scene coverage. A failed stage receives at most one bounded revision under the default configuration. For the text-grounded pipeline, inline shaders are rejected if they embed reference source code, leave unresolved Shadertoy interfaces, omit a fallback, or exceed their declared contract. After compaction and independent validation of query fidelity and execution freedom, the implementation handoff supplies the task-specific query to the coding agent. The three approved planning outputs are used to construct that query rather than concatenated into its execution request. The downstream process records the development trajectory and final game artifact, which are admitted under the checks in Appendix~\ref{app:react_quality_checks}.

\end{document}